\documentclass[10pt,twocolumn,letterpaper]{article}

\usepackage{iccv}
\usepackage{times}
\usepackage{epsfig}
\usepackage{graphicx}
\usepackage{amsmath}
\usepackage{amssymb}
\usepackage{graphicx}
\usepackage{stfloats}
\usepackage{placeins}
\usepackage{multirow}
\usepackage{enumitem}
\usepackage{booktabs}
\usepackage{stmaryrd}
\usepackage{stfloats} 
\usepackage{diagbox}
\usepackage{makecell}
\usepackage[table]{xcolor}
\usepackage{tikz}
\usepackage{subcaption}
\usepackage{soul}
\usepackage{dsfont}

\usepackage[accsupp]{axessibility}

\usepackage[pagebackref=true,breaklinks=true,letterpaper=true,colorlinks,bookmarks=false]{hyperref}
\newtheorem{definition}{Definition}

\usepackage[capitalize]{cleveref}
\crefname{section}{Sec.}{Secs.}
\Crefname{section}{Section}{Sections}
\Crefname{table}{Table}{Tables}
\crefname{table}{Tab.}{Tabs.}

\iccvfinalcopy

\ificcvfinal\fi

\begin{document}

\title{\vspace{-1em}Convolutional Networks with Oriented 1D Kernels }

\author{Alexandre Kirchmeyer
\thanks{Work done during an internship at Princeton University.}
\\
Carnegie Mellon University\\
{\tt\small akirchme@cs.cmu.edu}
\and
Jia Deng\\
Princeton University\\
{\tt\small jiadeng@princeton.edu}
}

\maketitle
\ificcvfinal\thispagestyle{empty}\fi

\begin{abstract}
In computer vision, 2D convolution is arguably the most important operation performed by a ConvNet. Unsurprisingly, it has been the focus of intense software and hardware optimization and enjoys highly efficient implementations. 
In this work, we ask an intriguing question: can we make a ConvNet work without 2D convolutions? Surprisingly, we find that the answer is yes---we show that a ConvNet consisting entirely of \emph{1D convolutions} can do just as well as 2D on ImageNet classification. Specifically, we find that one key ingredient to a high-performing 1D ConvNet is \emph{oriented 1D kernels}: 1D kernels that are oriented not just horizontally or vertically, but also at other angles.
Our experiments show that oriented 1D convolutions can not only  replace 2D convolutions but also augment existing architectures with large kernels, leading to improved accuracy with minimal FLOPs increase. 
A key contribution of this work is a highly-optimized custom CUDA implementation of oriented 1D kernels, specialized to the depthwise convolution setting. Our benchmarks demonstrate that our custom CUDA implementation almost perfectly realizes the theoretical advantage of 1D convolution: it is faster than a native \emph{horizontal} convolution for any \emph{arbitrary angle}. Code is available at \url{https://github.com/princeton-vl/Oriented1D}.  
\end{abstract}

\vspace{-1em}

\section{Introduction}
\label{sec:intro}

\vspace{-0.5em}

Convolutional Networks (ConvNets) \cite{lecun_convolutional_1998, krizhevsky_imagenet_2012} are widely used in computer vision. They have been successfully applied to a variety of tasks \cite{girshick_fast_2015, redmon_you_2016, guo_segnext_2022, zheng_clrnet_2022} and domains \cite{krizhevsky_imagenet_2012, martinez_beyond_2017, irvin_chexpert_2019, ravanbakhsh_estimating_2017, tajbakhsh_convolutional_2016}, and many new ConvNet-based building blocks \cite{szegedy_going_2014, dai_deformable_2017, xie_aggregated_2017} and design practices \cite{sandler_mobilenetv2_2019, tan_efficientnet_2020, liu_convnet_2022}  have emerged over the years.

In the computer vision context, a 2D convolution is arguably the most important operation performed by a ConvNet. In virtually all ConvNet architectures \cite{krizhevsky_imagenet_2012, he_deep_2015, liu_convnet_2022}, 2D convolution is the default choice and accounts for the bulk of the computation. Unsurprisingly, 2D convolution has been the focus of intense software and hardware optimization and enjoys highly efficient implementations. 

In this work, we ask an intriguing  question: can we make a ConvNet work without 2D convolution? Surprisingly, we find that the answer is yes---we show that a ConvNet consisting entirely of \emph{1D convolutions} can do as well on ImageNet classification, a surprising result given that 2D convolution has been the go-to design choice. 

Specifically, we find that one key ingredient to a high-performing 1D ConvNet is \emph{oriented 1D kernels}: 1D kernels that are oriented not just horizontally or vertically, but also at other angles. This is a novel finding---although horizontal and vertical 1D convolutions have been frequently used in the past,  1D kernels oriented at arbitrary angles have not been well studied. 

Oriented 1D kernels are motivated by the fact that 2D kernels can be approximated by 1D kernels, which are more efficient computationally. In particular, it is well known that convolution with a separable 2D kernel (i.e.\@ rank 1 as a matrix) is equivalent to consecutive convolutions with a vertical 1D kernel and a horizontal 1D kernel, leading to significant efficiency gain. However, in practice, not all learned 2D kernels are rank 1; if we only use vertical and horizontal 1D kernels, 2D kernels with a full rank, such as diagonal matrices, are poorly approximated, which can lead to a loss in accuracy: for example, the network may be less able to detect a 45$^\circ$ edge. This is when oriented 1D kernels can be helpful. By orienting a 1D kernel at more angles, we expand the set of 2D kernels that can be well approximated by 1D kernels while retaining the efficiency of 1D kernels. 

Oriented 1D kernels are also motivated by the increasing use of large 2D kernels in recent convolutional architectures. Large 2D kernels improve the modeling of long-range dependencies, which have been shown to result in better accuracy \cite{liu_convnet_2022, ding_scaling_2022, liu_more_2022}. However, large 2D kernels are significantly more expensive because the cost scales quadratically. A $31\times 31$ kernel is $19$ times more costly in terms of multiply-add operations than the standard $7\times 7$ kernels. This motivates oriented 1D kernels as an alternative for modeling long-range dependencies, because the cost of 1D kernels scale only linearly with the kernel size.  

The concept of oriented 1D kernels is simple, but non-trivial to implement in a way that realizes its efficiency advantage over 2D kernels. This is because on a GPU, the pattern of memory access matters as much as, if not more than, the number of floating point operations. Applying a 1D kernel oriented at an arbitrary angle requires accessing non-contiguous data; a naive implementation can easily negate the theoretical advantage of 1D kernels due to poor management of memory access. In addition, it is important for the implementation to not introduce significant memory overhead, which could be incurred by some naive implementations. Note that while horizontal/vertical 1D convolutions are well supported and optimized by existing libraries, 1D convolutions at an arbitrary angle is not. 

A key contribution of this work is a highly-optimized custom CUDA implementation of oriented 1D kernels, specialized to the setting of depthwise convolution \cite{chollet_xception_2017}, where each kernel is applied to only one depth channel. We optimize for depthwise convolution, because it has become an essential building block in recent state-of-art architectures \cite{tan_efficientnetv2_2021, liu_convnet_2022}, with superior accuracy-efficiency trade-offs. In addition, we find depthwise convolution to be the more useful setting in our experiments. Experiments show that our custom CUDA implementation almost perfectly realizes the theoretical advantage of 1D convolution: our 1D convolution at an \emph{arbitrary angle} is faster than the native \emph{horizontal} 1D convolution in PyTorch, which is highly optimized and achieves over 96\% of the theoretical speedup over 2D convolution. Notably, our implementation incurs minimal memory overhead; it uses less than 5\% more GPU memory than the native horizontal 1D convolution in PyTorch. Our implementation is open-sourced as a plug-and-play PyTorch module at \url{https://github.com/princeton-vl/Oriented1D}. 

With our custom implementation, our experiments show that oriented 1D convolution  can not only  replace 2D convolution but also augment existing architectures with large kernels, leading to improved accuracy with minimal FLOPs increase. 
We expect our implementation to be a useful primitive for further innovation in neural architectures for computer vision. 

Our main contributions are two-fold. 
First, we present the novel finding that state-of-the-art accuracy can be achieved with oriented 1D convolution alone, and that oriented 1D convolution can be a useful primitive to augment existing architectures. 
Second, we introduce an optimized CUDA implementation of depthwise oriented 1D convolution that nearly maxes out the theoretical efficiency of 1D convolution. 
\section{Related Work}

\vspace{-0.45em}

\paragraph{Modern ConvNets.} In recent years we have witnessed the emergence of novel training techniques \cite{liu_convnet_2022, tan_efficientnet_2020, tan_efficientnetv2_2021} and block designs \cite{chollet_xception_2017, hu_squeeze-and-excitation_2019, sandler_mobilenetv2_2019} inspired by transformers \cite{liu_convnet_2022} and neural architecture search \cite{tan_efficientnet_2020}. Depthwise convolutions \cite{chollet_xception_2017} have become essential components of many ConvNets \cite{sandler_mobilenetv2_2019, tan_efficientnet_2020, liu_convnet_2022} for their superior accuracy/computation trade-off. In search of better accuracy/computation trade-offs, recent research has looked at better scaling \cite{tan_efficientnetv2_2021, xie_aggregated_2017}, fused operations \cite{tan_efficientnetv2_2021, ding_scaling_2022}, neural architectural search \cite{tan_efficientnet_2020} and sparsity \cite{liu_more_2022} amongst others. Our work fits into this search for better efficiency by looking at 1D kernels that have the potential to scale better with kernel size. 

\vspace{-1.5em}

\paragraph{1D convolutions. } 
The use of 1D kernels has previously been explored in the context of image representation learning by \cite{peng_large_2017, szegedy_rethinking_2015, szegedy_inception-v4_2016,chen_xsepconv_2020}. These approaches rely primarily on decomposing a 2D convolution into a horizontal and vertical convolution \cite{rigamonti_learning_2013}, and are applied in ConvNets in a parallel \cite{peng_large_2017, guo_segnext_2022} or stacked design \cite{szegedy_inception-v4_2016}. However the use of 1D kernels often results in a drop in accuracy \cite{chen_xsepconv_2020}. Some approaches explore how to overcome this drawback, through SVD decomposition \cite{denton_exploiting_2014}, or L2 reconstruction loss \cite{jaderberg_speeding_2014}. 
Our approach tries to address this problem differently: we generalize 1D kernels to allow for non-horizontal and non-vertical kernels, whilst preserving the linear cost and advantages of 1D kernels. A recent work \cite{li_diagonal-kernel_2021} has explored a similar idea using diagonal kernels, but they only studied \emph{replacing} horizontal/vertical kernels with diagonal/anti-diagonal kernels, and reported better accuracy than horizontal/vertical kernels but worse accuracy than the 2D baseline. In contrast, we study 1D kernels oriented at more varied angles \emph{including} horizontal and vertical, while achieving no worse accuracy than the 2D baseline. Moreover, their work implements diagonal/anti-diagonal kernels by masking 2D kernels, thus gaining no efficiency advantage over 2D kernels, whereas we provide an optimized implementation that achieves near-perfect speedup. 

\vspace{-1.5em}

\paragraph{Oriented kernels. }
The use of oriented kernels is not new and dates back at least to steerable filters \cite{freeman_design_1991}. A commonly used approach is to learn a decomposition of a kernel on a basis which supports rotation, like harmonic functions \cite{worrall_harmonic_2017}, wavelet transforms \cite{oyallon_deep_2015} or steerable filters\cite{weiler_learning_2018}. The end goal is often to make the networks invariant or equivariant to rotation, and other transformations \cite{cohen_group_2016, weiler_learning_2018}. 
Our approach is simpler than methods using external kernel functions. This allows us to design fast implementations that can be integrated easily into existing architectures. Contrary to group equivariant \cite{cohen_group_2016} methods, we do not attempt to design rotation-equivariant networks but only to expand the expressiveness of 1D kernels with arbitrary orientations.

\vspace{-1.1em}

\paragraph{Large kernel design.} The emergence of ConvNets benefitting from large kernels \cite{liu_convnet_2022} has defied the long-established and implied superiority of small kernels since first initiated by VGGNet \cite{Simonyan15}. This has led to a wave of research exploring new ways to improve performance with bigger kernel sizes. RepLKNet \cite{ding_scaling_2022} achieves better performance with 31$\times$31 kernels by integrating a small 3$\times$3 or 5$\times$5 kernel that compensates for the difficulty in learning large kernels. SlaK \cite{liu_more_2022} pushes this principle further with rectangular 5$\times$51 and 51$\times$5 kernels combined with dynamic sparsity. Other lines of research include dilated convolutions \cite{yu_multi-scale_2016}, deformable convolutions \cite{dai_deformable_2017, zhu_deformable_2018} and continuous convolutions \cite{romero_ckconv_2021, romero_towards_2022}.
Our approach differentiates itself from these past works by studying large oriented 1D kernels, a type of large kernels that offers unique efficiency advantages but have not been studied in the context of modern ConvNet architectures. 
\vspace{-0.5em}
\section{Oriented 1D kernels}
\vspace{-0.25em}
In this section we introduce oriented 1D kernels for depthwise convolution, provide a computational analysis to justify their use and design models to test their effectiveness
\vspace{-1em}
\subsection{Definition}
\begin{figure}[b]
\vspace{-1em}
    \centering
    \small
    \includegraphics[width=0.9\linewidth]{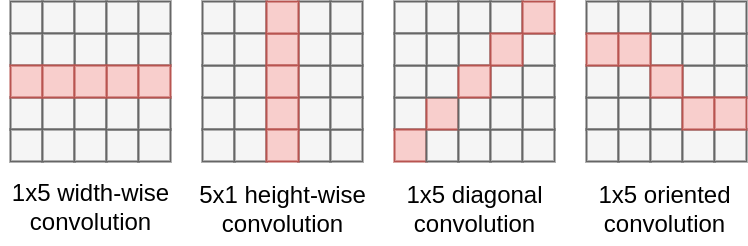}
    \vspace{-1em}
    \caption{Examples of oriented 1D kernels}
    \label{fig:illustration}
    \vspace{-0.5em}
\end{figure}

\vspace{-0.25em}
An oriented kernel is a kernel which is not always applied along the horizontal or vertical axis but along any axis as specified by an angle/direction $\theta$. We illustrate this concept in \Cref{fig:illustration}. For more expressivity, we allow $\theta$ to change at every channel, and call this per-channel angle $\boldsymbol \theta_c$. \Cref{def:def} translates this intuition into a mathematical formulation. A justification can be found in the appendix.

\vspace{-0.5em}
\begin{definition}[Depthwise convolution of oriented 1D kernel]
    \label{def:def}
    Let $\mathbf{x} \!\in\! \mathbb{R}^{N\times H \times W \times C}$ be the input of a depthwise convolution by an oriented 1D kernel of weights $\mathbf{w} \in \mathbb{R}^{K \times C}$ and per-channel angles $\boldsymbol \theta \in \mathbb{R}^C$. We define the output $\mathbf{y} \in \mathbb{R}^{N\times P \times Q \times C}$ as: $\forall n\!\in\!\llbracket0, N\!-\!1\rrbracket, p\!\in\!\llbracket0, P\!-\!1\rrbracket, q\!\in\!\llbracket0, Q\!-\!1\rrbracket, c\!\in\!\llbracket0, C\!-\!1\rrbracket,$\vspace{-0.5em}
    \begin{align} 
        & \mathbf{y}_{npqc} = \sum_{k=0}^{K-1} \mathbf{x}_{nhwc} \mathbf{w}_{kc}
        \label{eqn:definition} \\
        &\text{where} \begin{cases} h = str\cdot p + \left \lfloor -(k-pad) \cdot \sin \boldsymbol \theta_c \right \rfloor\\ 
        w = str\cdot q + \left \lfloor \text{ }(k-pad) \cdot \cos \boldsymbol \theta_c \right \rfloor \end{cases}\label{eqn:coordinate}
    \end{align} 
    $N$ is the batch size, $C$ is the number of input channels, $K$ is the 1D kernel size, $H,W$ and $P,Q$ are input and output dimensions, $str$ is the stride and $pad$ is the padding.
\end{definition}

The use of rounding in \Cref{eqn:coordinate} is necessary because of the discrete nature of input and filter coordinates. As an alternative to rounding down the filter offsets, bilinear interpolation could be considered. However in our experience, interpolation increases the computational cost substantially, making it impractical. See appendix for more details.

In this paper we only consider fixed $\boldsymbol \theta$, meaning that $\boldsymbol \theta$ is not learnt. Instead, we pick a fixed number $D$ of angles $0, \frac{1}{D}180^\circ, \frac{2}{D}180^\circ ..., \frac{D-1}{D}180^\circ$ and partition the channels into $D$ equal groups such that angle $i$ is associated to group $i$. In other words, for every channel of group $i$, it is assigned to angle $\frac{i}{D} 180^\circ$. In practice, we consider $D \in \{2, 4, 8\}$ and the case $D=C$. For $D = 4$ and $C = 512$, this means that there are 4 groups of 128 channels, with respective angles $0, 45^\circ, 90^\circ, 135^\circ$. We will call $D$ the number of directions as it is more visual.

\subsection{Computational cost analysis}

Depthwise convolutions are nearly always used in conjunction with pointwise convolutions with normalizations \cite{ba_layer_2016,ioffe_batch_2015} and/or non-linearities \cite{agarap_deep_2019, hendrycks_gaussian_2020} added in between. The idea is that depthwise convolutions mix spatial information and pointwise convolutions mix channel information \cite{haase_rethinking_2020}.
The combination of both is called a depthwise separable convolution (DSC) \cite{chen_xsepconv_2020, sandler_mobilenetv2_2019} and allows depthwise convolutions to be used as 2D convolution approximators. 

In this subsection, we show that switching to oriented 1D kernels leads to substantial speedups in computing both depthwise convolutions and DSCs. 

To demonstrate this, notice that a DSC with $C'$ output channels is the mix of a depthwise convolution and of a pointwise convolution, with respective computational costs of $NCHWK^2$ and $NCHWC'$ Multiply-Adds (MADs). Normalizing by the input size $NCHW$, a DSC induces a cost of: 

\vspace{-1em}$$\frac{NCHWK^2 + NCHWC'}{NCHW} = K^2 + C'\vspace{-0.25em}$$
By replacing the 2D kernel with a 1D kernel, the expression reduces to:
\vspace{-0.25em}$$\frac{NCHWK + NCHWC'}{NCHW} = K + C'\vspace{-0.25em}$$

According to our benchmarks \Cref{table:computational_analysis} on an NVIDIA RTX 3090 with Pytorch 1.11, pointwise convolutions are at least $20\times$ more efficient than depthwise convolutions. As such, depthwise convolutions account for more than 50\% of the workload when $K^2 \geq C'/20$. For typical ConvNets, $C' \sim 10^ 3$, therefore the expression simplifies to $K \approx 7$, which is the kernel size used by ConvNeXt. This implies that we can expect a significant gain by replacing 2D kernels by 1D kernels on a network like ConvNeXt. This observation provides the basis for our study of oriented 1D kernels and motivates our search for fast implementations.

\begin{table*}[t]

    \centering
    \small{
    \begin{tabular}{c|cccccccc} 
     \hline
     Variant & MADs & Kernel & Inference & Training & MADs & Runtime & Efficiency & Memory \\
     &  & Size $K$ & Runtime & Runtime & Ratio & Ratio &  & Usage\\
     \hline
     Convolution {\scriptsize (PyTorch)} & $C'K^2$ & 7 & 34.0$\pm$0.2ms & 78.4$\pm$0.3ms & & & & 9.7G \\
     \hline 
     Depthwise 2D {\scriptsize (PyTorch)} & $K^2$ & 7 & 8.3$\pm$0.1ms & 22.8$\pm$0.1ms & \multirow{2}{2em}{0.096} & \multirow{2}{1em}{2.3} & \multirow{2}{1.75em}{24.3} & 0.77G \\
     Pointwise {\scriptsize (PyTorch)} & $C'$ & 1 & 4.2$\pm$0.1ms & 9.8$\pm$0.1ms &  &  &  & 9.7G\\
     \hline
     Depthwise 2D {\scriptsize (PyTorch)} & $K^2$ & 7 & 8.3$\pm$0.1ms & 22.8$\pm$0.1ms & \multirow{2}{2em}{1.58} & \multirow{2}{1.6em}{1.53} & \multirow{2}{1.8em}{0.97} & 0.77G \\
     Depthwise 1D {\scriptsize (PyTorch)} & $K$ & 31 & 5.2$\pm$0.1ms & 14.9$\pm$0.3ms & & & & 0.77G\\
     \hline
     Depthwise 1D {\scriptsize (PyTorch)} & $K$ & 31 & 5.2$\pm$0.1ms & 14.9$\pm$0.3ms & \multirow{2}{0.5em}{1} & \multirow{2}{1.6em}{1.51} & \multirow{2}{1.9em}{1.51} & 0.77G\\
     Oriented Depthwise 1D  {\scriptsize (Ours)} & $K$ & 31 & \textbf{4.2$\pm$0.1ms$^1$} & \textbf{9.9$\pm$0.2ms$^1$} & & &  & 0.77G \\
     \hline
     \footnotesize{2D Depthwise Separable Conv. (DSC)} & $K^2 + C'$ & 7 & 12.7$\pm$0.1ms & 37.4$\pm$0.2ms & \multirow{2}{1.5em}{1.03} & \multirow{2}{1.6em}{1.56} & \multirow{2}{1.9em}{1.49} & 2.3G\\
     Oriented 1D DSC  {\scriptsize (Ours)} & $K + C'$ & 31 & \textbf{8.4$\pm$0.1ms} & \textbf{23.8m$\pm$0.2ms} & & & & 2.3G\\ \hline 
    \end{tabular}}
\vspace{-0.8em}
\caption{\small \textbf{Pairwise comparisons of the practical and theoretical performances of 2D and 1D convolutions.}  Oriented 1D kernels lead to significant speedups in theory and practice. We benchmark on a $56 \times 56$ input with FP32 on an NVIDIA RTX3090, against PyTorch\cite{pytorch} 1.11/CuDNN 8.2\cite{cudnn} with $N\!=\!64$ and $C\!=\!C'\!=\!512$. We compute the mean and standard deviation for 100 runs, preceded by 10 dry runs.   \textit{MADs} (Multiply-Adds) are divided by the input size $NCHW$ for better readability. \textit{Inference Runtime} measures only forward pass, \textit{Training Runtime} includes backpropagation. \textit{Memory Usage} measures peak memory usage difference before/after. Here, \textit{Efficiency} means \textit{Runtime Ratio} (practical gains) divided by \textit{MADs Ratio} (theoretical gains), and measures how efficient an implementation is compared to theory (higher is better). For $^1$, we aggregate runs over all integer angles $0^\circ, ..., 359^\circ$ to demonstrate that our implementation is fast for all angles. }
\vspace{-1.5em}
\label{table:computational_analysis}
\end{table*}

\Cref{table:computational_analysis} provides a summary of the theoretical and practical speedups that we get from switching to oriented 1D kernels. 
First, we observe that replacing the 2D 7$\times$7 kernel with the 1D 1$\times$31 kernel reduces MADs by 58\% even though we also switch to a larger kernel size. 
Second, we see that 1D kernels can be as efficient as 2D kernels: on PyTorch, switching to 1D leads to a 53\% practical runtime improvement which closely matches the theoretical 58\% MAD improvement. Third, our implementation is more efficient than PyTorch:  even though it supports oriented kernels, it is up to 51\% faster for all angles. Finally, we demonstrate that oriented 1D leads to significant overall speedups, as replacing the 2D 7$\times$7 DSC with the oriented 1D $K=31$ DSC reduces runtime \underline{as a whole} by $35\%$. 

\vspace{-0.5em}

\section{Model Instantiation}

\vspace{-0.25em}

To demonstrate the effectiveness of oriented 1D kernels, we gradually improve a ConvNeXt \cite{liu_convnet_2022} architecture with 1D kernels. 
We propose 3 different models: \textit{ConvNeXt1D}, \textit{ConvNeXt1D++} and \textit{ConvNeXt2D++} to show that oriented 1D ConvNets can be made as accurate as 2D ConvNets, and that we can use oriented 1D kernels to improve the accuracy of existing ConvNets. We structure our findings as follows: 1) \textit{Transition from 2D to 1D}, 2) \textit{1D-augmented design}, 2) \textit{1D/2D-mixed design}. 

\vspace{-0.25em}

\subsection{ConvNeXt reference}

\vspace{-0.25em}

Our models all use ConvNeXt \cite{liu_convnet_2022} depicted in \Cref{fig:convnext} as base architecture, which can be decomposed into: 

\begin{enumerate}[leftmargin=\parindent,align=left,labelwidth=\parindent,labelsep=0pt]
    \vspace{-0.75em}
    \item a \textit{stem layer}, which downsamples an input image 4$\times$ before feeding it to the rest of the network. It constitutes what we call the \textit{stem design}.  
    \vspace{-0.5em}
    \item 4 \textit{stages} each composed of a series of ConvNext blocks as defined by the \textit{block design}. They are responsible for the bulk of the ConvNet and are typically very deep \cite{he_deep_2015}.
    \vspace{-1.5em}
    \item A \textit{downsampling layer} separating each \textit{stage}, inducing a hierarchical structure that defines ConvNets \cite{Simonyan15, he_deep_2015}. 
    
\end{enumerate}
\begin{figure}[h]
    \centering
    \includegraphics[width=\linewidth]{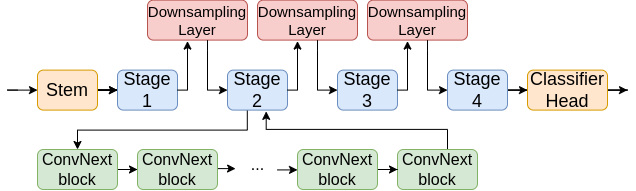}
    \vspace{-2em}
    \caption{Illustration of ConvNeXt which acts as the base architecture for all of our models.}
    \vspace{-1.5em}
    \label{fig:convnext}
\end{figure}

\noindent We define the ``\textit{2D Block} design"  and ``\textit{2D Stem} design" after ConvNeXt's block and stem designs and present them in \Cref{fig:stem_convnext} and \Cref{fig:block_2d}. In short, the ``\textit{2D Block} design" used by ConvNeXt is an inverted bottleneck with depthwise convolutions \cite{chollet_xception_2017}, as proposed initially by MobileNetV2 \cite{sandler_mobilenetv2_2019}. It integrates GeLU \cite{hendrycks_gaussian_2020} and LayerNorm \cite{ba_layer_2016}, following similar adoption in transformers \cite{devlin_bert_2019, liu_swin_2021}. 
ConvNeXt uses an aggressive downsampling strategy for its stem layer in the form of a 4$\times$4 stride 4 convolution. 
ConvNeXt further innovates by using an even-sized 2$\times 2$ stride 2 convolution as downsampling layer. Even-sized kernels tend to degrade performance because they introduce asymmetric padding \cite{zhang_making_2019, wu_convolution_2019}. In our experience, modifying this layer leads to a drop in accuracy, which aligns with observations made in \cite{liu_convnet_2022}. Consequently, we keep the original downsampling layer in our models. 

In the following, we consider the ConvNeXt architecture and progressively improve the stem design and block design, using oriented 1D kernels. \Cref{table:summary_models} presents a summary of our models and \Cref{fig:block_stem_design} the designs that we use.

\vspace{-0.5em}

\begin{table}[h]
    \centering
    \small
    \begin{tabular}{l|cc} 
     \hline
     Model & Stem Design & Block Design \\
     \hline
     ConvNeXt & 2D & 2D \\
     ConvNeXt1D & Depthwise 1D & 1D \\
     ConvNeXt1D++ & Depthwise 1D & 1D++ \\
     ConvNeXt2D++ & 2D & 2D++ \\
     \hline
    \end{tabular}
\vspace{-0.5em}
\footnotesize
\caption{\textbf{Model Summary.} We use ConvNeXt as baseline and propose \textbf{ConvNeXt1D}, a fully 1D ConvNet, \textbf{ConvNeXt1D++}, a 1D-augmented ConvNet, \textbf{ConvNeXt2D++}, a mixed 1D/2D ConvNet.\looseness=-1 }
\vspace{-1.5em}
\label{table:summary_models}
\end{table}

\subsection{Transition from 2D to 1D}

\begin{figure*}[t]
\vspace{-0em}
     \centering
     \begin{subfigure}[b]{0.1\linewidth}
         \centering
         \includegraphics[scale=0.28]{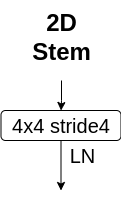}
         \caption{}
         \label{fig:stem_convnext}
     \end{subfigure}
     \hfill
     \begin{subfigure}[b]{0.1\linewidth}
         \centering
         \includegraphics[scale=0.28]{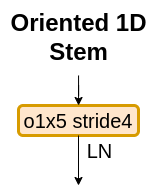}
         \caption{}
         \label{fig:stem_oriented}
     \end{subfigure}
     \hfill
     \begin{subfigure}[b]{0.1\linewidth}
         \centering
         \includegraphics[scale=0.28]{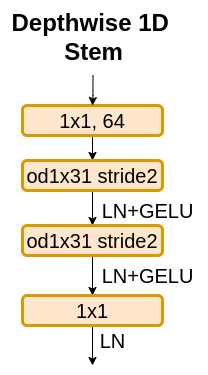}
         \caption{}
         \label{fig:stem_depthwise}
     \end{subfigure}
     \hfill
     \begin{subfigure}[b]{0.1\linewidth}
         \centering
         \includegraphics[scale=0.28]{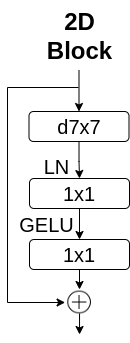}
         \caption{}
         \label{fig:block_2d}
     \end{subfigure}
     \hfill
     \begin{subfigure}[b]{0.1\linewidth}
         \centering
         \includegraphics[scale=0.28]{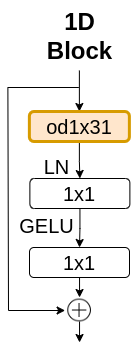}
         \caption{}
         \label{fig:block_1d}
     \end{subfigure}
     \hfill
     \begin{subfigure}[b]{0.1\linewidth}
         \centering
         \includegraphics[scale=0.28]{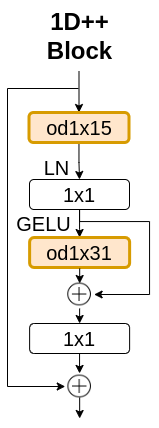}
         \caption{}
         \label{fig:block_1dpp}
     \end{subfigure}
     \hfill
     \begin{subfigure}[b]{0.1\linewidth}
         \centering
         \includegraphics[scale=0.28]{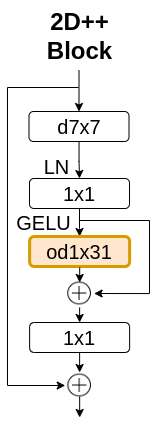}
         \caption{}
         \label{fig:block_2dpp}
     \end{subfigure}
     \vspace{-0.75em}
    \caption{\textbf{Stem and block designs} used in our models. The 2D \textbf{ConvNeXt} baseline is a combination of \textbf{(a)+(d)}, our fully 1D \textbf{ConvNeXt1D} network is a mix of \textbf{(c)+(e)}, our 1D-augmented \textbf{ConvNeXt1D++} model is composed of \textbf{(c)+(f)} and \textbf{ConvNeXt2D++}, our mixed 1D/2D network, by \textbf{(a)+(g)}. Stem \textbf{(b)} is used as baseline for our ablation \Cref{table:oriented}. \textit{o} stands for oriented, \textit{d} for depthwise. }
    \vspace{-1.5em}
    \label{fig:block_stem_design}
\end{figure*}

\paragraph{Oriented 1D network.} As a starting point, we construct a simple 1D baseline which we progressively improve to get \textit{ConvNeXt1D}. It is obtained by taking a ConvNeXt architecture and doing a one-on-one translation of every 2D kernel to an oriented 1D kernel. The network is defined through its ``\textit{1D Stem} design" and ``\textit{1D Block} design" as shown in \Cref{fig:stem_oriented} and \Cref{fig:block_1d}, which are just the oriented 1D equivalents of the \textit{2D Stem/Block} designs. Notice that the \textit{1D Stem} is a 1$\times$5 kernel, which is done to avoid even-sized kernels \cite{wu_convolution_2019}. We use this network to evaluate our oriented 1D models in  \Cref{table:oriented}.

\vspace{-1.25em}

\paragraph{Depthwise oriented 1D Stem Design. }
\label{par:depthwise} Naively transitioning from 2D to 1D as defined above naturally leads to performance degradation as the number of parameters in the stem layer is reduced by more than 65\%. To account for this, we propose a new stem design better suited for 1D networks, dubbed the \textit{Depthwise 1D Stem}. It is inspired by RepLKNet \cite{ding_scaling_2022} and is shown in \Cref{fig:stem_depthwise}. \textit{Depthwise 1D Stem} is a series of depthwise and pointwise convolutions: the stem alternates between expanding and mixing channels, or downsampling the input and mixing spatial information. For this stem design, we introduce the parameter $C_0$, which represents the number of channels by which we initially expand the input. In ablation \Cref{table:stem} we show that choosing $C_0$ large enough is necessary to preserve the accuracy of the model. This suggests that the number of spatial parameters is a key component in a ConvNet's accuracy.

\vspace{-1.25em}

\paragraph{Large kernels. } Oriented 1D kernels are very advantageous from a long-range modeling perspective. Because their cost is linear in $K$, kernels of size 1$\times$31 are cheaper than 2D kernels of size 7$\times$7. The interest of having such large $K$s is that the network is able to achieve global scale convolutions very early/deep in the network, in fact as soon as the 2$^\text{nd}$ stage. This stems from the observation that the input to stage 2 will be of size 28$\times$28 because of the downsampling operated at the stem layer and end of stage 1. As $31 > 28$, this means that $K$ is bigger than the input size, allowing it to encode all long-range dependencies at stage 2 and use the extra depth to model more complex spatial interactions. Note that we limit per-stage $K$ to $[31,31,27,15]$ or twice the stage input size, to avoid uninitialized weights. 

\vspace{-1.25em}

\paragraph{Layer-wise rotation.} We can improve the spatial mixing of oriented 1D kernels by adding an angular shift at every layer. We call this \textit{layer-wise rotation}. By adding a layer-wise rotation of 90$^\circ$ at every layer, we can reproduce a horizontal kernel in one layer and a vertical kernel in the next layer, effectively approximating a 2D kernel in a 2-layer setup. We integrate this idea in all of our models to improve performance, as validated by ablation \Cref{table:layerwise}.

\vspace{-1.25em}

\paragraph{Downsampling layer.} We claim that we can make a 2D network fully 1D without actually changing 2$\times$2 downsampling layers. This is based on the observation that a 2$\times$2 kernel can be seen as the sum of a diagonal and anti-diagonal kernel, which are special cases of oriented 1D kernels. See more details in the appendix. 

\vspace{-1.25em}

\paragraph{ConvNeXt1D.} We can now combine all of these findings to define our fully oriented 1D network \textit{ConvNeXt1D}. It is constructed on top of ConvNeXt with the \textit{Depthwise 1D Stem} and \textit{1D Block} defined above, and uses a large kernel of size $K=31$ with $D=8$ directions, as suggested by our ablation study \Cref{table:ablation_kd}. We show in \Cref{sec:experiments} that \textit{ConvNeXt1D} is able to perform on par with ConvNeXt.

\vspace{-0.5em}

\subsection{1D-Augmented Block Design}

\vspace{-0.25em}

With \textit{ConvNeXt1D}, we have presented a 1D network that can compete with existing 2D networks. We now provide evidence that oriented 1D kernels can also improve the accuracy/computation trade-off of ConvNets by boosting performance for a negligible increase in FLOPS/parameters. 

For that purpose, we propose the \textit{ConvNeXt1D++} model, constructed on top of \textit{ConvNeXt1D}. \textit{ConvNeXt1D++} reuses the \textit{Depthwise 1D Stem} that defines \textit{ConvNeXt1D} and introduces the improved \textit{1D++ Block} design which builds upon \textit{1D Block} and is presented in \Cref{fig:block_1dpp}. The idea is to reduce $K\!=\!31$ to $K\!=\!15$ and insert into the \textit{1D Block} a depthwise convolution of a large oriented 1$\times$31 kernel. This depthwise convolution is added residually to the network. We specifically target the \textit{inverted bottleneck} layer as shown in \Cref{fig:block_1dpp}, where we expect long-range dependencies to help the most. The goal is to integrate global scale interactions into local representations. With this design, we obtain better performance compared to \textit{ConvNeXt}, even though the extra FLOPS/parameters represent less than 5\% of the total. See \Cref{sec:experiments} for more results.

\vspace{-0.25em}

\subsection{1D/2D-Mixed Block Design}

\vspace{-0.25em}

We now demonstrate the flexibility of our method by proposing the \textit{ConvNext2D++} model. This model is constructed by extending the 2D ConvNext architecture with a new block design, the \textit{2D++ Block}, as shown in \Cref{fig:block_2dpp}. The \textit{2D++ Block} design is constructed on top of the \textit{ConvNeXt Block} and adds to it the same 1D augmentations that were included in the \textit{1D++ Block}. Through this process, we show that oriented 1D kernels can be dropped in, \textit{as is}, in existing architectures to improve long-range modeling and accuracy. Similarly as for \textit{ConvNeXt1D++}, we get better performances than \textit{ConvNeXt}, demonstrating that our approach is general and can be applied successfully to state-of-the-art models. 
\section{Experiments}
\label{sec:experiments}

\vspace{-0.25em}

\noindent In this section, we present experimental results conducted on ImageNet \cite{deng_imagenet_nodate} image classification, and downstream taks including ADE20K \cite{ade20k} semantic segmentation and COCO \cite{coco} object detection.
We summarize our training setup below and present details in appendix.

\vspace{-0.5em}

\begin{table*}[b]
    \scriptsize
    \centering
    \begin{tabular}{l||cccc|ccc|cccc|cc}
     \hline
     & \multicolumn{4}{c|}{Image classification} & \multicolumn{3}{c|}{Semantic segmentation} & \multicolumn{4}{c|}{Object detection} & \multicolumn{2}{c}{Throughput (img/s)} \\
     Model & \#Params & FLOPs & Acc & EMA Acc & \#Params & FLOPs & mIoU & \#Params & FLOPs & AP$^\text{box}$ & AP$^\text{mask}$ & Ours & PyTorch \\
     \hline
     \rowcolor{gray!10} ConvNext-T \cite{liu_convnet_2022} & 28.6M & 4.5G & & 82.1 & 60M & 939G & 46.7 & 86M & 741G & 50.4 & 43.7 &  & 1020 \\
     \rowcolor{gray!10} SlaK-T \cite{liu_more_2022} & 30M & 5.0G & & \underline{82.5} & 65M & 936G & \underline{47.6}$^\dagger$ & && \textbf{51.3} & 44.3 &  & 450 \\
     ConvNext-T (reprod.) & 28.6M & 4.5G & 81.8 & 82.0 & 60M & 939G & 46.6 & 86M & 741G & 50.2 & 43.6 & & 1020\\
     ConvNext-T-1D & 28.5M & 4.4G & 82.2 & 82.2 & 60M & 927G & 45.2 & 86M & 739G & 50.3 & 43.6 & 900 & \st{800} 940 \\
     ConvNext-T-1D++ & 29.2M & 4.7G & \textbf{82.4} & \textbf{82.7} & 61M & 927G & 47.4 & 86M & 739G & \textbf{51.3} & \textbf{44.5} & 630 & \st{390} 650 \\
     ConvNext-T-2D++ & 29.2M & 4.8G & \underline{82.3} & \underline{82.5} & 61M & 939G & \textbf{48.1} & 87M & 741G & \underline{51.2} & \underline{44.4} & 670 & \st{410} 690\\
     \hline 
     \rowcolor{gray!10} ConvNext-B \cite{liu_convnet_2022} & 89M & 15.4G & & \underline{83.8} & 122M & 1170G & 49.9 & 146M & 964G & 52.7 & 45.6 & & 460 \\
     \rowcolor{gray!10} RepLKNet-31B \cite{ding_scaling_2022} & 79M & 15.3G & & 83.5 & 112M & 1170G & \underline{50.6} & 137M & 965G & 52.2 & 45.2 & & 340\\
     \rowcolor{gray!10} SlaK-B \cite{liu_more_2022} & 95M & 17.1G & & \textbf{84.0} & 135M & 1172G & 50.2$^\dagger$ & & & & & & 210\\
     ConvNext-B (reprod.) & 89M & 15.4G & \textbf{83.6} & 83.7 & 122M & 1170G & 49.4 & 146M & 964G & 52.4 & 45.2 & & 460\\
     ConvNext-B-1D & 88M & 15.3G & \textbf{83.6} & \underline{83.8} & 122M & 1150G & 49.4 & 145M & 960G & \underline{52.8} & \underline{45.7} & 450 & \st{430} 460 \\
     ConvNext-B-1D++ & 90M & 15.8G & \underline{83.5} & \underline{83.8} & 123M & 1149G & \textbf{50.7} & 147M & 960G & \textbf{52.9} & \textbf{46.0} & 290 & \st{200} 310 \\
     ConvNext-B-2D++ & 91M & 15.9G & \textbf{83.6} & \textbf{84.0} & 124M & 1170G & 50.2 & 148M & 964G & \textbf{52.9} & \underline{45.8} & 290 & \st{200} 300 \\
     \hline
    \end{tabular}
\footnotesize
\vspace{-0.5em}
\caption{\textbf{Experimental results on ImageNet image classification, ADE20K semantic segmentation using UperNet, and COCO object detection using Cascade Mask R-CNN.} Overall, our models are competitive with ConvNeXt and state-of-the-art ConvNets on all tasks. Note that for downstream tasks, we use non-EMA backbones which means that performance improvements are not attributable to increased classification accuracy. Backbones are pre-trained on ImageNet-1K. We report image classification results  with and without Exponential Moving Average (EMA). $^\dagger$ SlaK only provides single-scale segmentation results. Throughput measures the inference speed of our proposed models on 1 {\small NVIDIA RTX3090}, \textit{PyTorch} substitutes oriented kernels with PyTorch horizontal kernels. FLOPs are computed on input sizes $224^2$, $1280\times 800$ and $2048\times 512$ respectively. The crossed-out numbers are from the ICCV 2023 version of this paper and are sub-optimal because of the accidental use of \textit{.permute()} instead of \textit{.permute().contiguous()}; they are now corrected in this version. } 
\vspace{-2em}
\label{table:results}
\end{table*}

\vspace{-1em}

\paragraph{ImageNet Training.}We use the same training settings as ConvNeXt \cite{liu_convnet_2022} to train our models. In summary, we train for 300 epochs with 20 epochs of linear warm-up, using the AdamW optimizer \cite{loshchilov_decoupled_2019}, data augmentations and regularizations like RandAugment \cite{cubuk_randaugment_2019}, Mixup \cite{zhang_mixup_2018}, Random Erasing \cite{zhong_random_2020}, Cutmix \cite{yun_cutmix_2019}, Stochastic Depth \cite{huang_deep_2016} and Label Smoothing \cite{szegedy_rethinking_2016}. The learning rate follows a cosine decay schedule and is set initially to 0.004. We use a batch size of 4096 and weight decay of 0.05. 

\subsection{Training Setup}

\vspace{-0.5em}

\paragraph{Model sizes. }We evaluate our models using the Tiny and Base complexities defined by ConvNeXt \cite{liu_convnet_2022}, which are parameterized by the channel sizes $C = (C_1, C_2, C_3, C_4)$ and numbers of blocks $B = (B_1, B_2, B_3, B_4)$ for all 4 stages. For our models using the \textit{Depthwise 1D Stem} design, we also introduce the number of channels $C_0$ by which the stem layer initially expands the input, as described in \Cref{par:depthwise}. We name our model variants respectively \textit{ConvNeXt-T/B-1D, ConvNeXt-T/B-1D++} and \textit{ConvNeXt-T/B-2D++} and aggregate the configurations in  \Cref{table:model_size}.

\vspace{-0.25em}

\begin{table}[h]
\centering
\setlength\tabcolsep{4pt}
\begin{tabular}{r||ccccc|cccc}

 \hline
 Size & $C_0$ & $C_1$ & $C_2$ & $C_3$ & $C_4$ & $B_1$ & $B_2$ & $B_3$ & $B_4$ \\
 \hline
 Tiny/T & 64 & 96 & 192 & 384 & 768 & 3 & 3 & 9 & 3 \\
 Base/B & 64 & 128 & 256 & 512 & 1024 & 3 & 3 & 27 & 3 \\
 \hline
\end{tabular}
\vspace{-0.8em}
\caption{Model configurations, taken directly from ConvNeXt \cite{liu_convnet_2022}}
\vspace{-1em}
\label{table:model_size}
\end{table}

\subsection{Experimental Results} \label{image_classification}

\paragraph{Image classification. } \Cref{table:results} compares our models against state-of-the-art ConvNets like ConvNeXt \cite{liu_convnet_2022}, RepLKNet \cite{ding_scaling_2022} and SlaK\cite{liu_more_2022}. We see that both \textit{ConvNeXt1D} and \textit{ConvNeXt1D++} Tiny and Base perform on par with ConvNeXt, which shows that oriented 1D convolutions are able to perform just as well as 2D convolutions. Additionally, \textit{ConvNeXt-T/B-2D++} are able to beat ConvNeXt-T/B by +0.5\%/+0.3\% which demonstrates that oriented 1D kernels are versatile as stand-alone networks and as extensions to existing networks.

\vspace{-1em}

\paragraph{Semantic segmentation. } We fine-tune UperNet \cite{upernet} on the ADE20K \cite{ade20k} dataset. We follow the same setup as ConvNeXt, that is we preserve all parameters apart from the backbone. Similarly as ConvNeXt, we use model non-EMA weights and multi-scale training for 160k iterations with a 512 crop size. We present our semantic segmentation results in \Cref{table:results}. Overall, ConvNeXt-1D \textit{Tiny} lags behind ConvNeXt but \textit{Base} catches up to it. The augmented networks achieve higher mIoU than ConvNeXt-T/B by at least +0.8 / +0.8 mIoU. Note that for \textit{Base} models, non-EMA image classification accuracy does not change: because our networks do not use EMA backbones, this means that the IoU improvement is not attributable to better image classification, but to the choice of architecture, thereby validating the effectiveness of oriented kernels on \textit{Base} models.

\vspace{-1em}

\begin{table*}[b]
\scriptsize
\centering
\begin{tabular}{ r|c|cccccccc|cccccccc } 
 \hline
  & & \multicolumn{8}{c|}{$K=7$} & \multicolumn{8}{c}{$K = 31$} \\
 & $H,W$ & $\theta=$ 0 & 22.5 & 45 & 67.5 & 90 & 112.5 & 135 & 157.5 & $\theta=$ 0 & 22.5 & 45 & 67.5 & 90 & 112.5 & 135 & 157.5 \\
 \hline
 \rowcolor{gray!10} PyTorch/CuDNN & $14^2$ & \multicolumn{8}{c|}{\underline{0.4}} & \multicolumn{8}{c}{1.1}\\
 \rowcolor{gray!10} CUTLASS \cite{ding_scaling_2022} & $14^2$ & \multicolumn{8}{c|}{0.6} & \multicolumn{8}{c}{\textbf{0.6}}  \\
 \rowcolor{gray!10} Ours & $14^2$ & \textbf{0.3} & \textbf{0.3} & \textbf{0.3} & \textbf{0.3} & \textbf{0.3} & \textbf{0.3} & \textbf{0.3} & \textbf{0.3} & 0.8 & 0.8 & 0.8 & \underline{0.7} & \underline{0.7} & 0.8 & 0.8 & 0.8\\
 \hline
 PyTorch/CuDNN & $28^2$ & \multicolumn{8}{c|}{\underline{1.4}} & \multicolumn{8}{c}{3.8}\\
 CUTLASS \cite{ding_scaling_2022} & $28^2$ & \multicolumn{8}{c|}{2.3} & \multicolumn{8}{c}{\textbf{2.5}} \\
 Ours & $28^2$ & \textbf{0.9} & \textbf{0.9} & \textbf{0.9} & \textbf{0.9} & \textbf{1.0} & \textbf{0.9} & \textbf{0.9} & \textbf{0.9} & \underline{2.6} & \underline{2.6} & \underline{2.6} & \underline{2.6} & \textbf{2.5} & \underline{2.6} & \underline{2.6} & \underline{2.6}\\
 \hline
 \rowcolor{gray!10} PyTorch/CuDNN & $56^2$ & \multicolumn{8}{c|}{\underline{5.5}} & \multicolumn{8}{c}{15.0}\\
 \rowcolor{gray!10}CUTLASS \cite{ding_scaling_2022} & $56^2$ & \multicolumn{8}{c|}{9.4} & \multicolumn{8}{c}{\underline{10.2}} \\
 \rowcolor{gray!10} Ours & $56^2$ & \textbf{3.2} & \textbf{3.3} & \textbf{3.2} & \textbf{3.2} & \textbf{3.2} & \textbf{3.2} & \textbf{3.2} & \textbf{3.2} & \textbf{9.8} & \textbf{9.8} & \textbf{9.8} & \textbf{9.9} & \textbf{9.9} & \textbf{9.9} & \textbf{10.0} & \textbf{10.0}\\
 \hline
\end{tabular}
\vspace{-1em}
\caption{\textbf{Runtime comparison} of our oriented 1D implementation on 1 NVIDIA RTX 3090 for $N\!=\!64, C\!=\!512$, FP32. The mean is taken over 100 runs, preceded by 10 dry runs. We benchmark against the very competitive PyTorch/CuDNN and RepLKNet CUTLASS \cite{cutlass} implementations on horizontal convolutions. Our implementation outperforms PyTorch consistently regardless of angle $\theta$ thanks to intelligent data access patterns. }
\label{table:speed_1}
\end{table*}

\paragraph{Object detection. } We follow ConvNeXt and fine-tune a Cascade Mask R-CNN \cite{cascade_maskrcnn} on the COCO dataset using multi-scale training and a 3$\times$ schedule. As shown in \Cref{table:results}, ConvNeXt-1D achieves similar or better performances compared to ConvNeXt. The results confirm the expressiveness of augmented networks: they have +0.5 AP$^\text{box}$ and +0.6 AP$^\text{mask}$ compared to both RepLKNet\cite{ding_scaling_2022} and ConvNeXt. 

\paragraph{Inference speed metrics. } We complement our classification results with inference speeds on a $224\times 224$ input, batch size $64$, and 1 {\small NVIDIA RTX3090}, as presented in \Cref{table:results}. Even though 1D kernels are faster than 2D and the FLOPs increase is negligible, our ConvNeXt models are slower because of the modified Stem and Block designs. This suggests a model-level under-utilization due to memory related bottlenecks, which we will study as future work. To confirm our computational analysis, we replace oriented kernels with PyTorch 1D kernels. Note that we fuse certain 1D operations which improves 1D speed over 2D.

\smallskip \noindent \textbf{Additional architectures.} To provide evidence that our 1D convolutions are generalizable, we compare the 2D ConvNets MobileNetV3 \cite{mobilenetv3} and ConvMixer \cite{convmixer} against 1D equivalents, obtained by replacing every non-strided depthwise convolution with an oriented kernel. We train on ImageNet using the same setup as ConvNeXt. 1D models have 10\%-25\% higher inference throughput versus 2D models but lead to a drop of 0.8\%-1.1\% accuracy. Given that only substitution is done and no other tricks are used, this shows that 1D kernels show promise in working in other settings and architectures. Note that we target specifically an architecture which presents depthwise separable convolutions.

\vspace{+0.5em}

\resizebox{0.99\linewidth}{!}{
\hspace{-2.5em}
\begin{tabular}{l|cccccc} 
 \hline
 Model & Epochs & $K$ & \#Params & FLOPs & Throughput & Acc. \\ 
 \hline
 MobileNetv3-small 2D & 120 &  & 5.5M & 0.06G & 6800 {\footnotesize img/s} & \textbf{60.9} \\ 
 MobileNetv3-small 1D & 120 & & 5.5M & 0.06G &\textbf{7200 {\footnotesize img/s}} & 60.1 \\
 \hline
 ConvMixer-768/32 2D & 100 & 7 & 21.1M & 20.7G & 200{\footnotesize img/s} & \textbf{78.4}\\ 
 ConvMixer-768/32 1D & 100 & 15 & 20.3M & 20.1G & \textbf{250{\footnotesize img/s}} &  77.3 \\
 \hline
\end{tabular}}

\vspace{0.25em}

\subsection{CUDA kernel speed comparison}
\label{section:speed_analysis}

We compare the speed of our oriented 1D kernel implementation versus the competitive PyTorch and RepLKNet CUTLASS \cite{cutlass} implementations and aggregate the results in \Cref{table:speed_1}. We conduct the experiments on a single NVIDIA RTX 3090, using single-precision FP32. In some instances, our custom kernel outperforms the CuDNN/CUTLASS implementation by a substantial margin. These algorithms rely on the assumption that computation is the main bottleneck, thus their design focuses on maximizing computational throughput in a cache-friendly manner, at the expense of reading more data. As a result, they are slower for smaller kernel sizes where data efficiency is critical. This also makes them hard to adapt to oriented kernels because they assume linear access patterns which results in bad performance, as we show in appendix. 

\vspace{-0.5em}

\newpage 
\subsection{Ablation study}
\label{sec:ablation}

\vspace{-0.5em}

In this subsection, we carry out ablations to determine the influence on accuracy of characteristic oriented 1D kernel parameters. \Cref{table:oriented} compares oriented 1D models against non-oriented 2D baselines, 
\Cref{table:layerwise} studies the impact of layerwise rotation,
\Cref{table:ablation_kd} studies the best combination of kernel size and direction and 
\Cref{table:stem} looks at the best $C_0$. We comment the results in the next subsection.

\begin{table}[!h]
\footnotesize
\centering
\begin{tabular}{ ccccccc } 
 \hline
 Stem & Block & $K$ & $D$ & \#Params & FLOPs & Top-1 \\
 Design & Design & & & & &  Acc \\
 \hline
 \rowcolor{gray!25} 2D & 2D & 7$^2$ & & 28.6M & 4.5G & \textbf{82.0} \\
 \rowcolor{gray!25} 2D & 2D & 31$^2$ & & 30.0M & 5.8G & 81.6 \\
 Oriented 1D & 1D & 7 & 8 & 28.3M & 4.4G & 81.5 \\
 Oriented 1D & 1D & 31 & 8 & 28.5M & 4.4G & \underline{81.8}   \\ 
 \hline
 \rowcolor{gray!25} Depthwise 2D & 2D & 7$^2$ & & 28.6M & 3.9G & \underline{82.0} \\
 \rowcolor{gray!25} Depthwise 2D & 2D & 31$^2$ &  & 34.8M & 6.1G & 81.4\\
 Depthwise 1D & 1D & 7 & 8 & 28.3M & 4.4G & \underline{82.0} \\
 Depthwise 1D & 1D & 31 & 8 & 28.5M & 4.4G & \textbf{82.2} \\
 \hline
 \rowcolor{gray!25} Depthwise 2D & 2D+2D & 7$^2$+7$^2$ & & 29.9M & 4.3G & 82.1 \\
 Depthwise 1D & 1D++ & 15+31 & 8 & 29.2M & 4.7G & \textbf{82.7}\\
 2D & 2D++ & 7$^2$+31 & 8 & 29.4M & 4.7G & \underline{82.5} \\
 \hline
\end{tabular}
\vspace{-0.5em}
\footnotesize
\caption{\textbf{Comparison of oriented 1D models against 2D kernel baselines } on ConvNeXt-T. \textit{Depthwise 2D} is the 2D transposition of \textit{Depthwise 1D} and \textit{2D+2D} is the 2D transposition of both \textit{1D++} and \textit{2D++} block designs. Overall, by switching to 1D and changing the stem, we are able to consistently match the performance of 2D baselines. We discuss these results in \Cref{section:discussion}.}
\label{table:oriented}
\end{table}

\vspace{-1em}
\begin{table}[h]
\small
\centering
\begin{tabular}{ cccc } 
 \hline
 $K$ & $D$ & Layer-wise rotation & Top-1 Acc. \\
 \hline
 31 & 4 & $0^\circ$ & 82.01 \\
 31 & 4 & alternating $90^\circ$ & \textbf{82.11} \\
 \hline
\end{tabular}
\vspace{-0.5em}
\caption{\textbf{Layer-wise rotation} on ConvNeXt-1D. We see that adding layer-wise rotation increases accuracy. }
\label{table:layerwise}
\vspace{-0.5em}
\end{table}

\begin{table}[h]
\footnotesize
\centering
\begin{tabular}{ ccccc } 
 \hline
 $K$ & $D$ & \#Params & FLOPs & Top-1 Acc. \\
 \hline
 \rowcolor{gray!25} $7\times 7$ &  & 28.6M & 4.5G & \textbf{82.00} \\
 7 & 2 & 28.3M & 4.4G & 81.80 \\
 7 & 4 & 28.3M & 4.4G & \underline{81.90} \\
 7 & 8 & 28.3M & 4.4G & \textbf{82.00} \\
 7 & $C$ & 28.3M & 4.4G & 81.65 \\
 \hline
 15 & 2 & 28.4M & 4.4G & 81.84 \\
 15 & 4 & 28.4M & 4.4G & 82.05 \\
 15 & 8 & 28.4M & 4.4G & \textbf{82.14} \\
 15 & $C$ & 28.4M & 4.4G & \underline{82.12} \\
 \hline
 31 & 2 & 28.5M & 4.5G & 81.67 \\
 31 & 4 & 28.5M & 4.5G & \underline{82.01} \\
 31 & 8 & 28.5M & 4.5G & \textbf{82.16} \\
 31 & $C$ & 28.5M & 4.5G & 81.92 \\
 \hline
 63 & 2 & 28.5M & 4.5G & 81.73 \\
 63 & 4 & 28.5M & 4.5G & 81.92 \\
 63 & 8 & 28.5M & 4.5G & \textbf{82.11} \\
 63 & $C$ & 28.5M & 4.5G & \underline{81.97} \\
 \hline
\end{tabular}
\vspace{-0.5em}
\caption{\textbf{Optimal $K$ and $D$.} We compare ConvNext-T-1D with different kernel sizes $K$ and directions $D$. No layer-wise rotation is used. $D=8$ corresponds to the best number of directions and $K=31$ seems the most advantageous in that case. Having too few $D=2$ or too many directions $D=C$ seems to hurt performance.  }
\vspace{-0.5em}
\label{table:ablation_kd}
\end{table}

\begin{table}[h]
\small
\centering
\begin{tabular}{ cccccc } 
 \hline
 $K$ & $D$ & $C_0$ & \#Params & FLOPs & Top-1 Acc. \\
 \hline
 31 & 4 & 4 & 28.5M & 4.4G & 78.9 \\ 
 31 & 4 & 16 & 28.5M & 4.4G & \underline{81.67} \\ 
 31 & 4 & 64 & 28.5M & 4.4G & \textbf{82.01} \\ 
 \hline
\end{tabular}
\vspace{-0.25em}
\caption{\textbf{Optimal $C_0$. }We compare ConvNext-T-1D with different stem channel sizes $C_0$. $C_0$ is directly tied to the number of spatial parameters in the stem. Increasing $C_0$ does not change the number of parameters yet results in a significant improvement in accuracy. This means that the spatial mixing done by depthwise convolutions is critical in achieving good performance. We fix this value to 64 in our models. }
\label{table:stem}
\end{table}

\subsection{Discussion}
\label{section:discussion}

\vspace{-0.25em}

\paragraph{No loss in expressiveness.} 1D ConvNeXt  networks are \textit{consistently} able to perform on par or outperfom their 2D equivalents according to \Cref{table:oriented}. This shows that oriented 1D kernels can be made as expressive as 2D kernels provided they are integrated the right way. 

\vspace{-1em}

\paragraph{Depthwise 1D stem.} From \Cref{table:oriented}, we see that our proposed \textit{Depthwise 1D Stem} plays a key role in making 1D networks competitive with 2D networks, by reintroducing the stem spatial parameters that were lost switching to 1D.

\vspace{-1em}

\paragraph{Better long-range scaling with 1D.} \Cref{table:oriented} shows that 2D models are unable to benefit from large kernel sizes $K$. This contrasts with the 1D kernel case and implies that 1D kernels exhibit better long-range scaling than 2D kernels. 

\newpage

\section{Conclusion}

\vspace{-0.5em}

Our study suggests that oriented 1D kernels are viable alternatives to 2D kernels as they can be just as expressive, and are empirically better suited for larger kernel sizes. 

\paragraph{Acknowledgement} 
This work was partially supported by the
National Science Foundation under Award IIS-1942981.

\newpage
\appendix

\section*{Appendix}

In this appendix, we present additional experiments including ERF analysis in \Cref{sec:erf} and sparsity in \Cref{sec:sparsity} and present additional training plots in \Cref{training_plots}. 

We also provide a more theoretical overview of oriented 1D kernels, and describe how we come up with our formulation of oriented 1D kernels in \Cref{math_formulation}, discuss design choices of oriented 1D kernels in \Cref{design_choices}, prove that a 2$\times$2 downsampling layer can be seen as a sum of oriented kernels in \Cref{proof_downsampling}, outline how to learn orientation in \Cref{angle_backprop} and establish a connection with anisotropic gaussian kernels in \Cref{anisotropic_filters}. 

Finally, we provide implementation notes in \Cref{implementation_notes}, describe training settings in \Cref{training_settings} and outline limitations in \Cref{limitations}. 

\section{Effective Receptive Field (ERF) Analysis} \label{sec:erf}

\vspace{-0.5em}

\begin{figure}[b]
    \vspace{-1.5em}
     \centering
     \begin{subfigure}[b]{0.45\linewidth}
         \centering
         \includegraphics[scale=0.15]{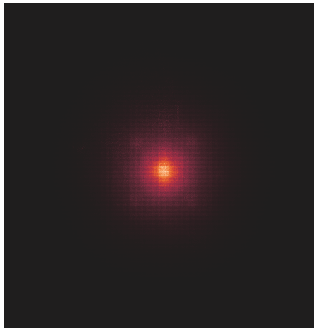}
         \vspace{-0.25em}
         \caption{2D}
         \vspace{-0.25em}
         \label{fig:erf_2d}
     \end{subfigure}
     \hfill
     \begin{subfigure}[b]{0.45\linewidth}
         \centering
         \includegraphics[scale=0.15]{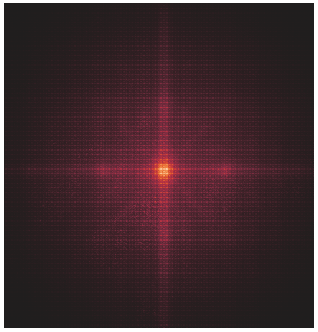}
         \vspace{-0.25em}
         \caption{1D}
         \vspace{-0.25em}
         \label{fig:erf_1d}
     \end{subfigure}
     
     \begin{subfigure}[b]{0.45\linewidth}
         \centering
         \includegraphics[scale=0.15]{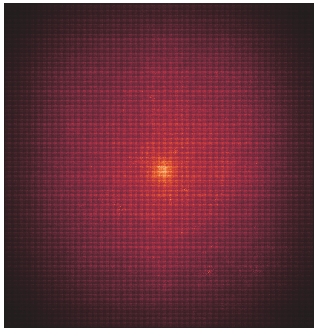}
         \vspace{-0.25em}
         \caption{2D++}
         \vspace{-0.25em}
         \label{fig:erf_2dpp}
     \end{subfigure}
     \hfill
     \begin{subfigure}[b]{0.45\linewidth}
         \centering
         \includegraphics[scale=0.15]{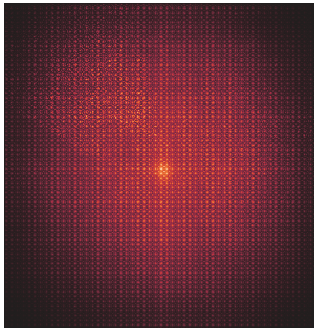}
         \vspace{-0.25em}
         \caption{1D++}
         \vspace{-0.25em}
         \label{fig:erf_1dpp}
     \end{subfigure}
     \vspace{-0.5em}
    \footnotesize
     \caption{Effective Receptive Field (ERF) for \textit{Base} models. A more widely distributed colored area indicates a larger ERF. We see that our networks increase the ERF significantly whilst preserving the overall shape. }
    \label{fig:erf}
\end{figure}

\begin{figure}[b]
    \vspace{-1.5em}
     \centering
     \begin{subfigure}[b]{\linewidth}
         \centering
         \includegraphics[scale=0.15]{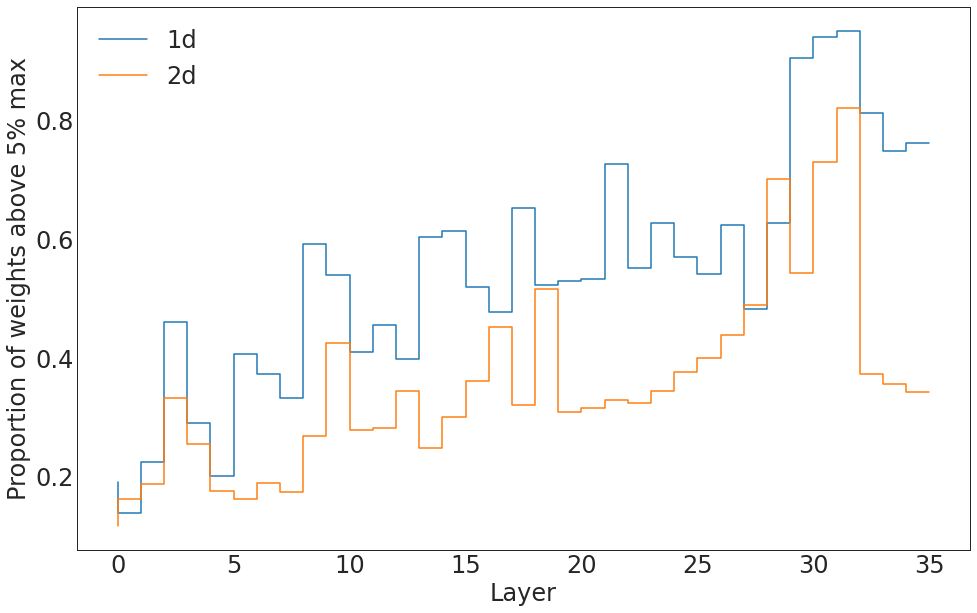}
         \vspace{-0.25em}
     \end{subfigure}
     \hfill
     \begin{subfigure}[b]{\linewidth}
         \centering
         \includegraphics[scale=0.15]{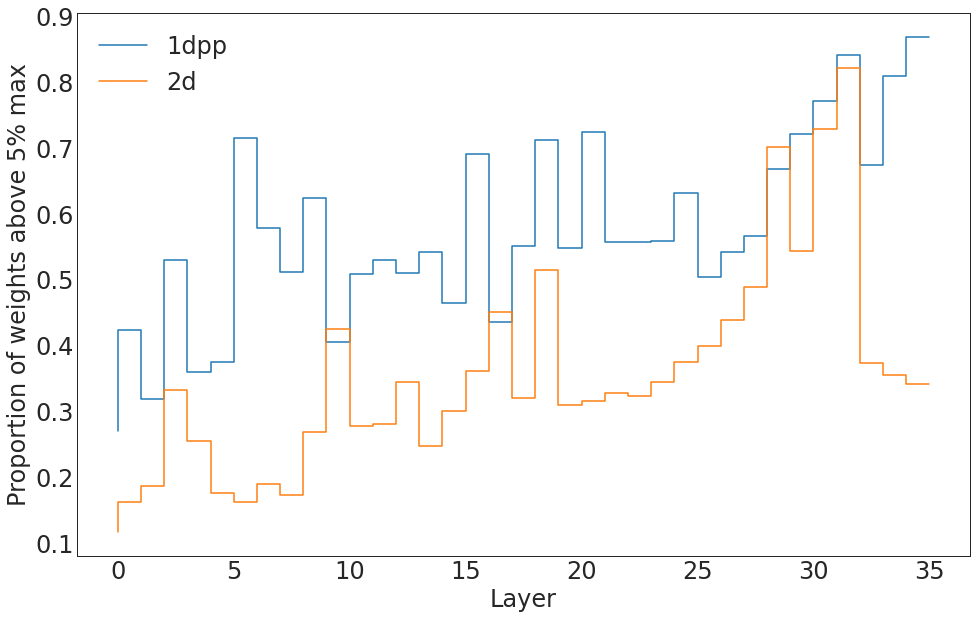}
         \vspace{-0.25em}
     \end{subfigure}
     \hfill
     \begin{subfigure}[b]{\linewidth}
         \centering
         \includegraphics[scale=0.15]{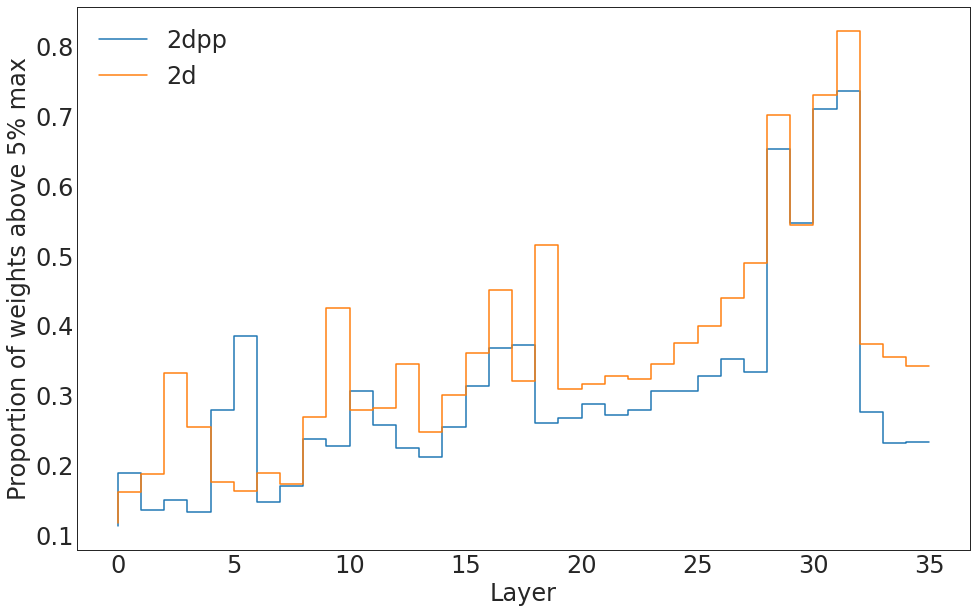}
     \end{subfigure}
     
     \vspace{-0.5em}
     \caption{Sparsity analysis on \textit{Base} models: Proportion of weights in absolute value above 5\% of maximum. We consider only the 1$^\text{st}$ depthwise convolution in each layer of stages 1-4. We see that fully 1D networks (1D and 1D++) are up to +20\% more dense than 2D networks (2D and 2D++). }
     \label{fig:sparse}
\end{figure}

In this section we provide an analysis of the ERF of our models.  According to \cite{luo2016}, ERFs scale in $O(K\sqrt{L})$ which is linear in the kernel size $K$ and only sub-linear in depth $L$. This demonstrates the advantage of kernel size over depth.

We follow the approaches provided by RepLKNet \cite{ding_scaling_2022} and SLaK \cite{liu_more_2022} to compute the ERF of our models. We sample 1000 images resized to $1024\times 1024$ from the ImageNet validation set, and for each pixel of every image we compute its contribution to the central point of the feature map generated in the last layer. We then average the contribution across all input channels and images. Results are shown in \Cref{fig:erf}. 

From \Cref{fig:erf}, we see that ConvNeXt has a concentrated ERF around the input center. This is different from ConvNeXt-1D, which has a significantly wider ERF. It manages to preserve the concentration around the center whilst widening the receptive field, albeit at the cost of introducing a horizontal/vertical bias. This bias is resolved in the augmented networks, which are able to preserve the concentration around the center and introducee a wide circular receptive field. Our networks are therefore able to balance the focus between local details and the attention to long-range dependencies. 

\vspace{-0.25em}

\section{Sparsity analysis} \label{sec:sparsity}

\vspace{-0.25em}
In this section, we analyze the sparsity introduced by the use of oriented 1D kernels in our 1D, 1D++ and 2D++ models. We consider only the 1$^\text{st}$ depthwise convolution in each layer of stages 1-4. This depthwise convolution is present and common to all block designs. From \Cref{fig:sparse} we see that fully 1D networks (1D and 1D++) are up to +20\% more dense than 2D networks (2D and 2D++). This is to be expected as we decrease the number of parameters from $7 \times 7$ to $1 \times 31$ by design. It also confirms the usefulness of oriented 1D kernels for introducing large kernels in ConvNets: not only do they widen the receptive field, but they also reduce the number of spatial parameters required to do so, thereby enabling more efficient learning.

 \begin{figure}[t]
     \centering
     \begin{subfigure}[b]{\linewidth}
         \centering
         \includegraphics[scale=0.25]{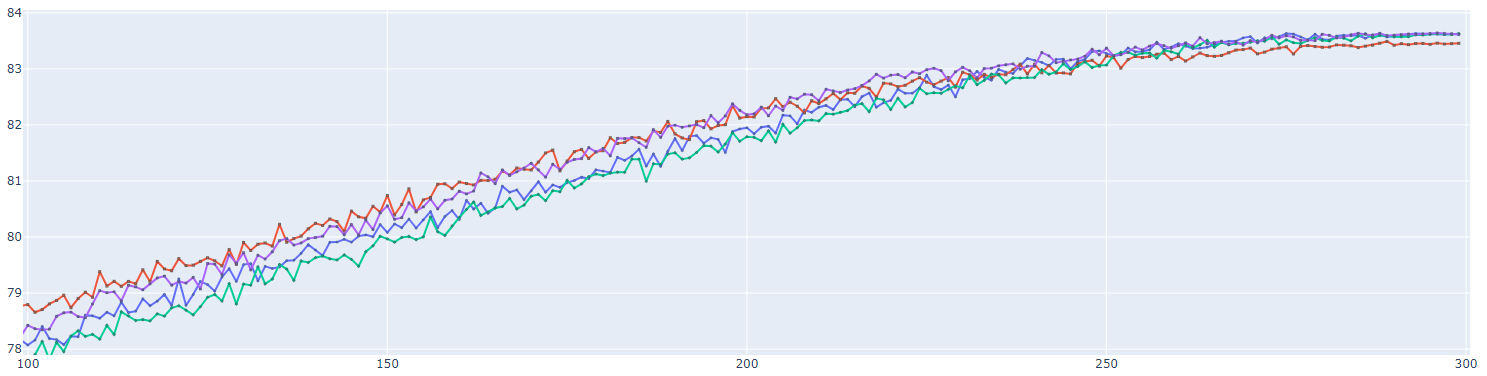}
         \caption{{\scriptsize no-EMA \color{blue}1D}/{\color{red}1D++}/{\color{green}2D}/{\color{violet}2D++}}
         \vspace{-0.25em}
         \label{fig:acc_non_ema}
     \end{subfigure}
     \hfill
     \begin{subfigure}[b]{\linewidth}
         \centering
         \includegraphics[scale=0.25]{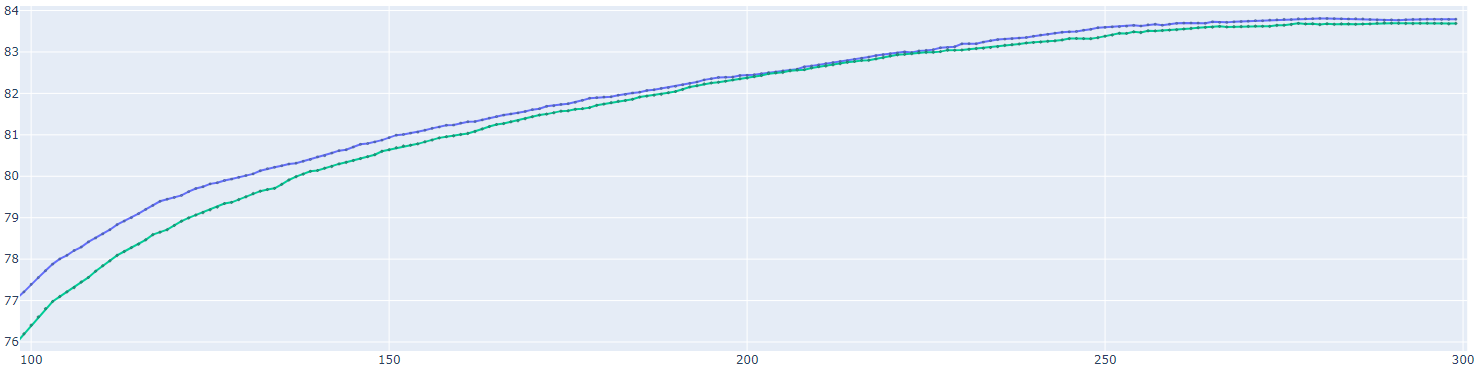}
         \caption{EMA {\color{blue}1D}/{\color{green}2D}}
         \vspace{-0.25em}
         \label{fig:acc_1d}
     \end{subfigure}
     
     \begin{subfigure}[b]{\linewidth}
         \centering
         \includegraphics[scale=0.25]{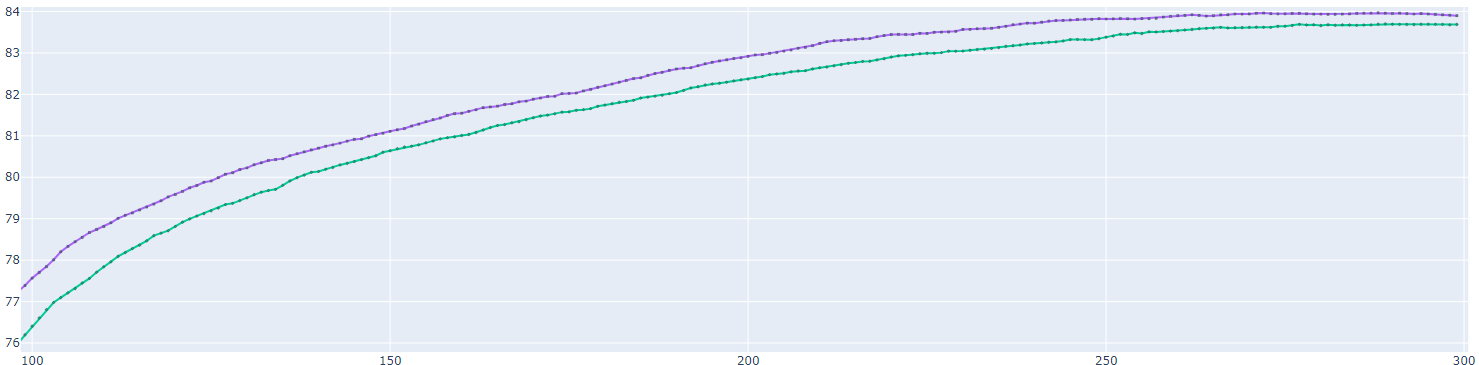}
         \caption{EMA {\color{violet}2D++}/{\color{green}2D}}
         \vspace{-0.25em}
         \label{fig:acc_2dpp}
     \end{subfigure}
     \hfill
     \begin{subfigure}[b]{\linewidth}
         \centering
         \includegraphics[scale=0.25]{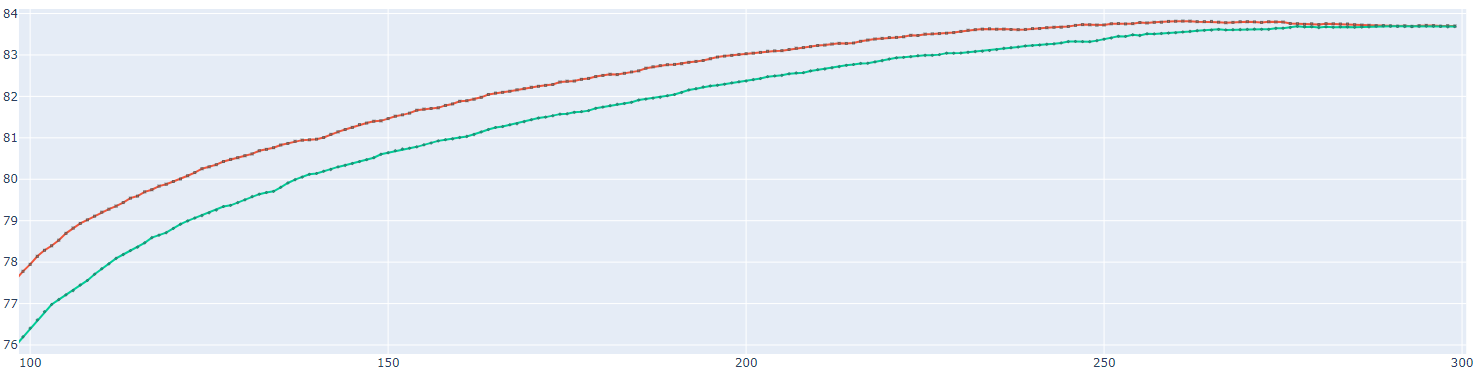}
         \caption{EMA {\color{red}1D++}/{\color{green}2D}}
         \vspace{-0.25em}
         \label{fig:acc_1dpp}
     \end{subfigure}
     \vspace{-1em}
    \footnotesize
     \caption{ImageNet EMA accuracy plots for \textit{Base} models. 2D in green, 2D++ in purple, 1D in blue, 1D++ in red. }
     \vspace{-1em}
    \label{fig:trajectories}
\end{figure}

\begin{figure}[t]
     \centering
     \begin{subfigure}[b]{\linewidth}
         \centering
         \includegraphics[scale=0.25]{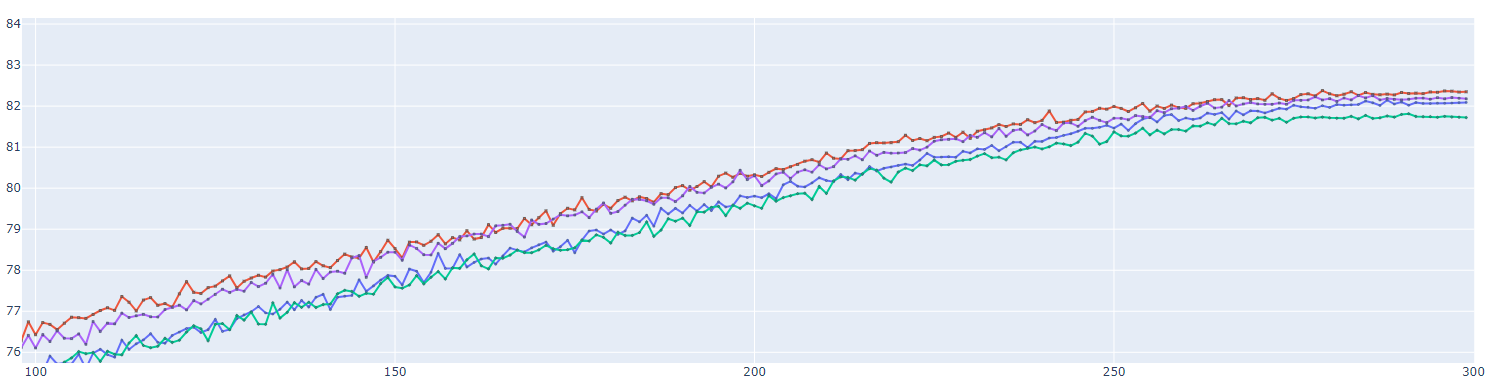}
         \caption{{\scriptsize no-EMA \color{blue}1D}/{\color{red}1D++}/{\color{green}2D}/{\color{violet}2D++}}
         \vspace{-0.25em}
         \label{fig:acc_non_ema_tiny}
     \end{subfigure}
     \hfill
     \begin{subfigure}[b]{\linewidth}
         \centering
         \includegraphics[scale=0.25]{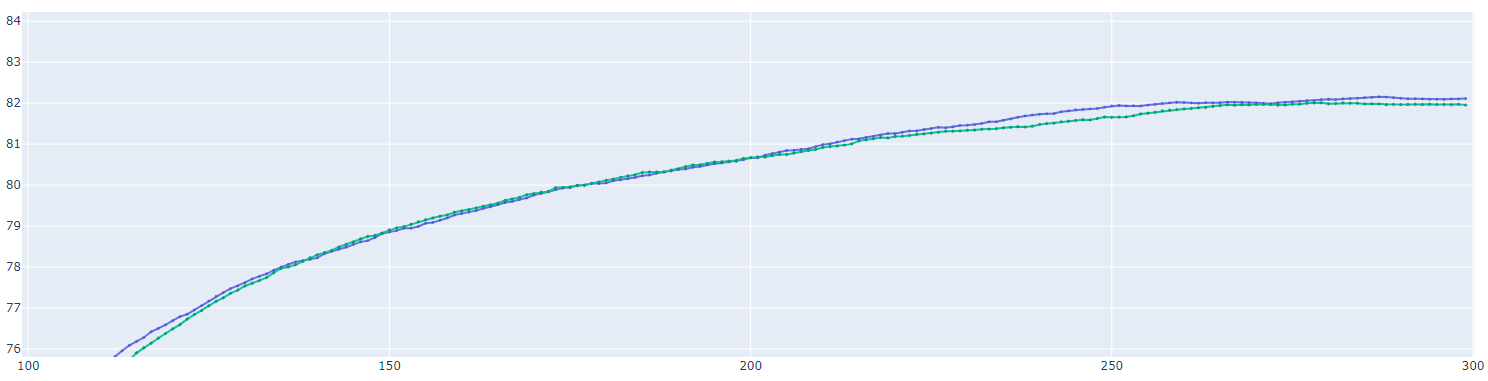}
         \caption{EMA {\color{blue}1D}/{\color{green}2D}}
         \vspace{-0.25em}
         \label{fig:acc_1d_tiny}
     \end{subfigure}
     
     \begin{subfigure}[b]{\linewidth}
         \centering
         \includegraphics[scale=0.25]{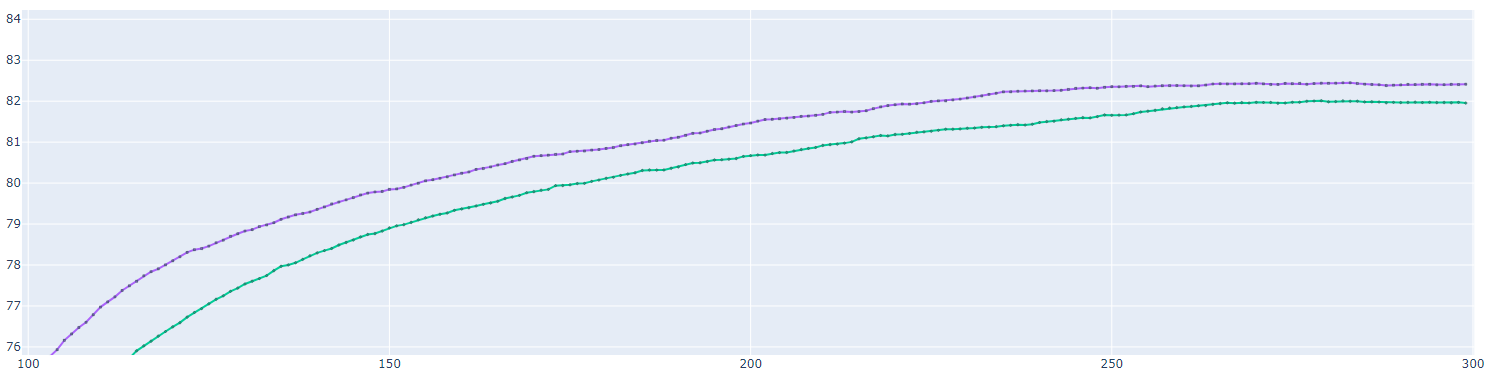}
         \caption{EMA {\color{violet}2D++}/{\color{green}2D}}
         \vspace{-0.25em}
         \label{fig:acc_2dpp_tiny}
     \end{subfigure}
     \hfill
     \begin{subfigure}[b]{\linewidth}
         \centering
         \includegraphics[scale=0.25]{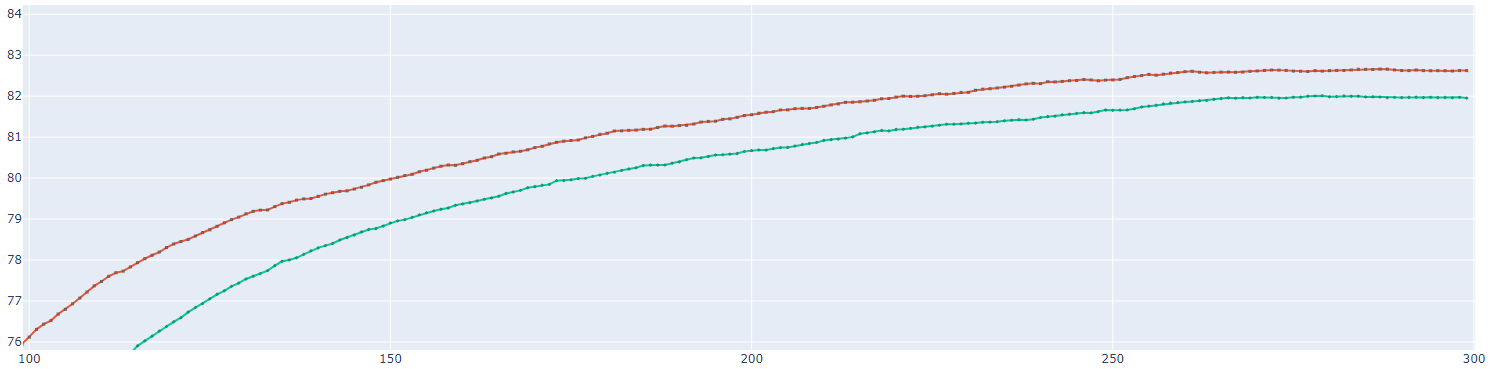}
         \caption{EMA {\color{red}1D++}/{\color{green}2D}}
         \vspace{-0.25em}
         \label{fig:acc_1dpp_tiny}
     \end{subfigure}
     \vspace{-1em}
    \footnotesize
     \caption{ImageNet EMA accuracy plots for \textit{Tiny} models. 2D in green, 2D++ in purple, 1D in blue, 1D++ in red. }
     \vspace{-1.5em}
    \label{fig:trajectories_tiny}
\end{figure}

\vspace{-0.5em}

\section{Training plots} \label{training_plots}

\vspace{-0.5em}

In this section, we plot ImageNet validation set accuracies of \textit{Base} models in \Cref{fig:trajectories}. We also aggregate statistics on these trajectories in \Cref{table:stats_trajectories}. We see that 1D++/2D++ models exhibit +0.6 and +0.5 EMA accuracy improvements versus ConvNeXt when averaged over epochs 150-250. This suggests that augmented 1D networks perform better than ConvNeXt during training. 

For the 1D++ \textit{Base} model, this accuracy improvement during training is significantly larger  compared to the +0.1 final accuracy difference versus ConvNeXt. This discrepancy suggests that our 1D++ model overfits and underperforms versus what it is expected to achieve. We have not fine-tuned any hyperparameter other than 1D parameters, to enable fair comparisons with ConvNeXt. We think that understanding the causes of this overfitting would enable 1D networks to perform better than they currently do on larger scales. 

\begin{table}[h]
\small
    \centering
    \begin{tabular}{l|cc|c}
         Model & {\scriptsize Acc. Difference} & {\scriptsize Acc. Difference} & {\scriptsize Discrepancy}\\ 
         & {\scriptsize Epochs 150-250} & {\scriptsize Final} & \\
         \hline 
         ConvNeXt-T & / & / & /\\
         ConvNeXt-T-1D & +0.0 & +0.2 & +0.2\\
         ConvNeXt-T-1D++ & +0.8 & +0.7 & -0.1\\
         ConvNeXt-T-2D++ & +0.7 & +0.5 & -0.2\\
         \hline
         ConvNeXt-B & / & / & /\\
         ConvNeXt-B-1D & +0.1 & +0.1 & 0.0\\
         ConvNeXt-B-1D++ & +0.6 & +0.1 & -0.5\\
         ConvNeXt-B-2D++ & +0.5 & +0.3 & -0.2\\
    \end{tabular}
    \vspace{-0.25em}
    \caption{Accuracy differences w.r.t ConvNeXt. On the left, averaged over epochs 150-250, on the right, accuracy at epoch 300. \textit{Discrepancy} measures the difference between these 2 quantities. We see that there is a strong discrepancy for ConvNeXt-B-1D++ -- suggesting that the model is underperforming. }
    \label{table:stats_trajectories}
\end{table}

\newpage

\phantom{.}

\newpage

\section{Oriented 1D kernels: Mathematical formulation}
\label{math_formulation}

\vspace{-0.5em} 

In this section, we present the intuition behind the mathematical formulation of oriented 1D depthwise convolutions. 
We start by introducing oriented kernels in the more general 2D setting, discuss the intuition behind the formula and specialize it to the 1D case.

To simplify the problem, we will only consider a single angle $\theta$ but the formulation can be easily generalized to support a per-channel angle $\boldsymbol \theta \in \mathbb{R}^C$ as done in the paper.

\subsection{2D formulation}

\vspace{-0.5em} 

Let $\mathbf{x} \in \mathbb{R}^{N \times H \times W \times C}$ denote the input, $\mathbf{w} \in \mathbb{R}^{R \times S \times C}$ the filter and $\mathbf{y} \in \mathbb{R}^{N \times P \times Q \times C}$ the output of the depthwise convolution. $N$ is the batch size, $C$ the number of channels, $H,W$ the input height and width, $P,Q$ the output height and width and $R, S$ the filter height and width. Let $(pad_h, pad_w)$ be the padding, $(str_h, str_w)$ the stride and $\theta$ the angle of the oriented kernel.

\begin{definition}
  We define a depthwise convolution of an oriented 2D kernel as: 

    $\forall n\!\in\!\llbracket0, N\!-\!1\rrbracket, p\!\in\!\llbracket0, P\!-\!1\rrbracket, q\!\in\!\llbracket0, Q\!-\!1\rrbracket, c\!\in\!\llbracket0, C\!-\!1\rrbracket,$\vspace{-0.5em}
\begin{align}
& y_{npqc} = \sum_{0 \leq r < R, 0 \leq s < S } x_{nhwc} w_{rsc} \\
&\text{where} \label{eqn:coordinate2d}  \begin{pmatrix}h \\ w\end{pmatrix} = \mathbf{R}_\theta \begin{pmatrix}r - pad_h \\ s - pad_w\end{pmatrix} + \begin{pmatrix}p\cdot str_h \\ q \cdot str_w\end{pmatrix}\\
&\text{and } \mathbf{R}_\theta = \begin{pmatrix}\cos \theta & -\sin \theta\\ \sin \theta & \cos \theta \end{pmatrix} \text{ is a rotation matrix of angle $\theta$}
\end{align}
\end{definition}
We refer to \Cref{eqn:coordinate2d} as the \textit{coordinate equation}. Because $h,w,p,q,r,s$ are all integers we need to introduce a discretization scheme for this formula to make sense. We will ignore this issue for now and discuss it in \Cref{discretization}.

\vspace{-1em} 

\paragraph{Intuition.} The goal of \Cref{eqn:coordinate2d} is to rotate the filter by an angle $\theta$, which is accomplished by introducing the rotation matrix $\mathbf{R}_\theta$.
Intuitively, every increment of $r$ or $s$ results in an increase by $(\cos \theta, \sin \theta)^T$ or $(-\sin \theta, \cos \theta)^T$ of the left-hand side of \Cref{eqn:coordinate2d} after rotation by $\mathbf{R}_\theta$. By doing so, we are effectively changing the direction of the increment of $r$ and $s$ from the vertical and horizontal axes to arbitrary oriented axes.

\vspace{-1em} 

\paragraph{Convolution origin.} The origin of the convolution is obtained by taking $r \!= s \!= 0$: we see that it is the same as the non-oriented case. This is guaranteed by our choice of \Cref{eqn:coordinate2d} and is not necessarily preserved by other formulations. 

\vspace{-0.25em} 

\subsection{Discretization and Interpolation} \label{discretization}
\vspace{-0.25em} 
\Cref{eqn:coordinate2d} makes the assumption that we can sample from a continuous input domain. However, due to the discrete nature of the input and filter, it becomes necessary to introduce a \textit{discretization} scheme. 

The simplest discretization scheme consists of rounding down the coordinates on the right-hand side of \Cref{eqn:coordinate2d}: this approach is fast but results in coarse approximations of orientation.

An alternative approach would be to do bilinear interpolation which allows us to account for finer angles. However, bilinear interpolation increases the MADs by a factor of 4$\times$ and results in at least 2$\times$ the runtime on our internal benchmarks \Cref{bilinear}. This makes an oriented 1$\times$31 kernel 2$\times$ as expensive as a 7$\times$7 kernel. 

In this paper, we choose to adopt the round-down discretization scheme, to showcase the usefulness and practicality of oriented 1D kernels. Under this scheme, \Cref{eqn:coordinate2d} becomes:

\vspace{-0.5em} 

\begin{align} \label{eqn:discret2d}
    \begin{pmatrix}h \\ w\end{pmatrix} = \left \lfloor \mathbf{R}_\theta \begin{pmatrix}r - pad_h \\ s - pad_w\end{pmatrix} \right \rfloor + \begin{pmatrix}p\cdot str_h \\ q \cdot str_w\end{pmatrix}
\end{align}
As discussed in this section, there exists many valid discretization schemes, and we leave their exploration as future work. 

\begin{table}[h]
\centering
\begin{tabular}{c|ccc}
Implementation & $K$ & Inference & Total \\
& & Runtime & Runtime \\
\hline
Round-down 1D & 31 & 4.2$\pm$0.1ms & 9.9$\pm$0.1ms\\
Bilinear 1D & 31 & 8.9$\pm$ 0.1ms  & 22.3$\pm$0.1ms \\
2D & 7 & 8.3$\pm$0.1ms  & 22.6$\pm$0.1ms
\end{tabular}
\caption{Runtime comparison between bilinear interpolation and other approaches. Bilinear interpolation results in a 2$\times$ speed reduction. The tests are done on an NVIDIA RTX 3090 for $N=64, C=512, H=W=56$, aggregated over 100 runs preceded by 10 dry runs. \textit{2D} refers to a PyTorch/CuDNN 2D depthwise convolution. \textit{Inference Time} measures only forward pass, \textit{Training Time} includes backpropagation. }
\label{bilinear}
\end{table}

\vspace{-1.5em} 

\subsection{1D formulation}

\vspace{-0.25em} 

Oriented depthwise 1D convolutions are specializations of \Cref{eqn:coordinate2d} for $r\!=pad_h\!=0$, $pad_w\!=pad$ and $str_h\!=str_w\!=str$. We use round-down as our discretization, as shown in \Cref{eqn:discret2d}, which results in the following equation:

\vspace{-1em} 

\begin{equation}
\begin{pmatrix}h \\ w \end{pmatrix} = str \cdot \begin{pmatrix}p \\ q\end{pmatrix} + \left \lfloor (k-pad) \cdot \begin{pmatrix}-\sin \theta \\ \cos \theta \end{pmatrix} \right \rfloor
\label{eqn:coordinate1d}
\end{equation}
where $K$ is the kernel size and $k$ varies from $0$ to $K-1$.
This is the formulation of oriented 1D kernels that we have considered in the paper. 

\textit{From now on, we will restrict our study to \Cref{eqn:coordinate1d}.}

\vspace{-0.25em} 

\section{Oriented 1D kernels: Design choices}
\label{design_choices}

\vspace{-0.25em} 

Let us now delve into the design choices regarding oriented 1D kernels.

\vspace{-0.25em} 

\subsection{Padding}
\label{apdx:padding}

\vspace{-0.25em} 

The way we express padding in \Cref{eqn:coordinate1d} turns out to be essential in preserving accuracy. If we were to naively replicate the non-oriented case and write \Cref{eqn:coordinate1d} as:

\vspace{-1em} 
\begin{equation}
\begin{pmatrix}h \\ w \end{pmatrix} = str \cdot \begin{pmatrix}p \\ q\end{pmatrix} + \left \lfloor k \cdot \begin{pmatrix}-\sin \theta \\ \cos \theta \end{pmatrix} \right \rfloor - pad
\label{eqn:padding}
\end{equation}

then the resulting formula would lead to different behaviors for vertical and horizontal convolutions and more generally result in uncentered convolutions. In turn, this greatly degrades accuracy. In contrast, \Cref{eqn:coordinate1d} adapts naturally to different angles and for $pad = \left \lfloor \frac{K}{2} \right \rfloor$ does not result in accuracy degradation. We consider only odd $K$ in our experiments and fix $pad = \left \lfloor \frac{K}{2} \right \rfloor$ to obtain centered oriented convolutions.

\vspace{-0.25em} 

\subsection{Rotation vs Shearing}
\vspace{-0.25em} 

Instead of applying a rotation to the coordinates of the convolution we can consider more relevant transformations. Let's first rewrite \Cref{eqn:coordinate1d} as: 
\vspace{-1em} 

\begin{align}
& \begin{pmatrix}h \\ w \end{pmatrix} = str \cdot \begin{pmatrix}p \\ q\end{pmatrix} + \left \lfloor \begin{pmatrix}\delta h^k \\ \delta w^k \end{pmatrix} \right \rfloor
\end{align}
\vspace{-2em} 

\begin{align}\text{\hspace{-0.5em}in terms of the \textit{filter offsets}: } 
\begin{pmatrix}\delta h^k \\ \delta w^k\end{pmatrix} = \mathbf{R}_\theta \begin{pmatrix}0 \\ k-pad\end{pmatrix}
\end{align}

 \vspace{-0.5em}

Intuitively, the \textit{filter offset} $(\delta h^k, \delta w^k)^T$ is the offset in the input grid where we apply the filter weight $w_k$, relative to the convolution origin.

\begin{figure}[h!]
\centering
\begin{subfigure}{.45\linewidth}
 \centering
 \includegraphics[width=0.45\linewidth]{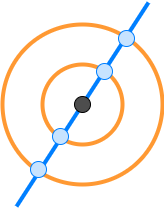}
 \vspace{-0.5em}
 \caption{Rotation }
 \vspace{-0.5em}
 \label{fig:rot}
\end{subfigure}
\begin{subfigure}{.45\linewidth}
 \centering
 \includegraphics[width=0.45\linewidth]{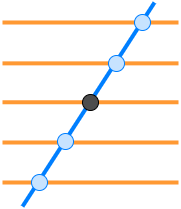}
 \vspace{-0.5em}
 \caption{Shear}
 \vspace{-0.5em}
 \label{fig:proj}
\end{subfigure}
\vspace{-0.5em}
\caption{Different Parameterizations}
\label{fig:rot_vs_proj}
\end{figure}

Instead of sampling \textit{filter offsets} on concentric circles with increasing and regular $k$ radii, we can choose to sample points lying on integer rows or columns. The first approach is equivalent to \textit{rotating} the filter axis as shown in \Cref{fig:rot}, whereas the second can be seen as \textit{shearing} the filter axis parallel to the columns or rows as depicted in \Cref{fig:proj}. 

The usefulness of formulation \ref{fig:proj} becomes clear on the following example. For $\theta\!=\!-45^\circ$, $pad\!=\!0$ and $k$ varying between $0$ and $K-1$, formulation \ref{fig:rot} outputs non-integer filter offsets $(\delta h^k, \delta w^k)^T = k(\sqrt{2}/2, \sqrt{2}/2)^T$, whereas formulation \ref{fig:proj} outputs integer filter offsets $(\delta h^k, \delta w^k)^T = k(1, 1)^T$. In this particular case, the latter approach removes the need for discretization.

More generally, formulation \ref{fig:rot} results in filter offsets $(-k\sin \theta, k\cos \theta)^T$. With rounding down as discretization, this is sub-optimal as it leads to some level of redundancy in filter offsets. 

Formulation \ref{fig:proj} is more complicated to handle and gives rise to two separate cases. If we intersect the filter axis with integer columns we get filter offsets $(-k\tan \theta, k)^T$. If we instead intersect with integer rows, we get offsets $(k, -k\cot \theta)^T$. By forcing one coordinate to be an integer we remove the redundancy encountered in formulation \ref{fig:rot} but we enlarge the kernel ``length" and introduce higher complexity due to the separate handling of cases.

Mathematically we can introduce the \textit{shear matrices} $\mathbf{S}^x_\theta = \begin{pmatrix}1 & -\tan \theta \\ 0 & 1\end{pmatrix}$ and $\mathbf{S}^y_\theta = \begin{pmatrix}1 & 0 \\ \cot \theta & 1\end{pmatrix}$ and rewrite filter offsets in terms of $\mathbf{S}^x_\theta$ and $\mathbf{S}^y_\theta$ instead of $\mathbf{R}_\theta$.

More thorough testing is necessary to compare both formulations, and is left for future work.

\vspace{-0.25em}

\section{Proof that the downsampling layer is the sum of 2 oriented kernels}
\label{proof_downsampling}

\vspace{-0.25em}

In this section, we prove the claim that we can express a downsampling layer as the sum of 2 oriented kernel convolutions by using an even-sized kernel \textit{specialization} of \Cref{eqn:coordinate1d} and carefully choosing the padding.

\vspace{-0.5em} 

\paragraph{Oriented convolution. } First, we define a vanilla convolution with an oriented 1D kernel as:

\vspace{-1em}

\begin{align}
& y_{npqg} = \sum_{0 \leq c < C, 0 \leq r < R, 0 \leq s < S } x_{nhwc} w_{grsc}
\end{align}

\vspace{-0.25em}

where $g$ denotes the output channel and $(h,w)^T$ verifies the \textit{coordinate} \Cref{eqn:coordinate1d}. Note that we suppose $\boldsymbol{\theta} = \theta_{gc}$ so that the angle can vary both with respect to the output and input channels. 

\vspace{-0.5em}

\paragraph{Even-sized kernel.} Next, we generalize \Cref{eqn:coordinate1d} for even-sized kernels by introducing an exterior padding $(pad_h, pad_w)$ which takes care of the padding asymmetry introduced by even-sized kernels \cite{wu_convolution_2019}.

\vspace{-1em}

\begin{equation}
\begin{pmatrix}h \\ w \end{pmatrix} = str \cdot \begin{pmatrix}p \\ q\end{pmatrix} + \left \lfloor (k-pad) \cdot \begin{pmatrix}-\sin \theta \\ \cos \theta \end{pmatrix} \right \rfloor - \begin{pmatrix}pad_h \\ pad_w \end{pmatrix}
\label{eqn:coordinate1d_even}
\end{equation}

\vspace{-1em}

\paragraph{Proof.} \underline{Claim 1}: We claim that a 2$\times$2 downsampling layer can be decomposed as the sum of a diagonal and anti-diagonal convolution. We use the following fact: if $W_1$ and $W_2$ are two 2D kernels and $I$ is any input, then the convolution of $I$ by $W=W_1+W_2$ equals the sum of the convolutions of $I$ by $W_1$ and convolution of $I$ by $W_2$. By defining $W_1$ as the 2$\times$2 diagonal kernel, and $W_2$ as the anti-diagonal kernel, this implies that the sum  $W=W_1+W_2$ can express any 2$\times$2 kernel, which proves our claim.

\underline{Claim 2}: We now claim that both diagonal and anti-diagonal kernels can be seen as even-sized oriented 1D kernels as defined in \Cref{eqn:coordinate1d_even}. We do this by carefully choosing $\theta$ and paddings $pad, pad_h$ and $pad_w$.
In fact, a diagonal kernel can be expressed using $\theta = -\frac{\pi}{4}, pad = 1-\sqrt{2}$ and $pad_h = pad_w = 0$.
Similarly, an anti-diagonal kernel can be expressed using $\theta = \frac{\pi}{4}, pad = 1-\sqrt{2}, pad_h = 1$ and $pad_w = 0$. Contrary to the diagonal kernel, we see that the anti-diagonal kernel requires the introduction of a vertical padding $pad_h = 1$. As explained earlier, this is necessary because of the asymmetry of even-sized kernels.

\underline{Summary:} We can now combine claims 1 and 2 to deduce that a downsampling layer can be seen as the sum of 2 oriented kernel convolutions. 

In this section, we have introduced even-sized oriented kernels. We leave their exploration as future work.

\vspace{-0.25em} 

\section{Angle backpropagation} \label{angle_backprop}
\vspace{-0.25em} 

Instead of seeing $\theta$ as a fixed parameter (we assume in this section that the same $\theta$ applies for all channels), we can instead try to learn it. To accomplish this, it becomes necessary to introduce a formulation of \Cref{eqn:coordinate1d} which is differentiable w.r.t $\theta$. 

The first idea is to replace the sum over the kernel size $k$ as a sum over $(h,w,k)$ and introduce a soft distribution $\omega_{pqhwk}(\theta)$ to weight the sum, as such:

\begin{equation}
\mathbf{y}_{npqc} = \sum_{h,w,k} \omega_{pqhwk}(\theta)\mathbf{x}_{nhwc}\mathbf{w}_{ck}
\label{eq:sum_omega}
\end{equation}
Here $h$ and $w$ are bound variables. Contrast with the original formulation where $h$ and $w$ are functions of $k$ as provided by \Cref{eqn:coordinate1d} 
\begin{equation}
\mathbf{y}_{npqc} = \sum_{k} \mathbf{x}_{nhwc}\mathbf{w}_{ck}
\label{eq:sum_noomega}
\end{equation}

\begin{figure}[h]
\centering
\begin{subfigure}{0.33\linewidth}
 \centering
 \includegraphics[width=\linewidth]{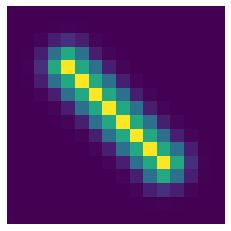}
 \caption{$\sigma=1.0$ }
\end{subfigure}
\begin{subfigure}{0.33\linewidth}
 \centering
 \includegraphics[width=\linewidth]{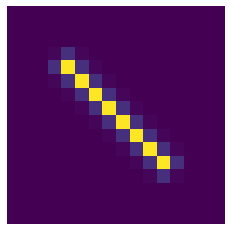}
 \caption{$\sigma=0.5$}
\end{subfigure}
\begin{subfigure}{0.33\linewidth}
 \centering
 \includegraphics[width=\linewidth]{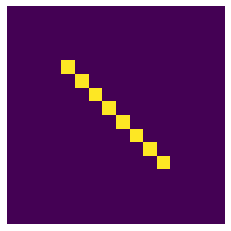}
 \caption{$\sigma=0.25$}
\end{subfigure}
\caption{Example of soft distribution for a diagonal kernel}
\label{fig:widths}
\end{figure}

The second idea is to relax the \Cref{eqn:coordinate1d} constraint to include non-zero contributions for $(h,w,k)$ which do not verify \Cref{eqn:coordinate1d} but are close to it. A natural choice is to model $\omega_{pqhwk}(\theta)$ as a gaussian distribution with variance $\sigma^2$ over the squared \Cref{eqn:coordinate1d} error, as illustrated in \Cref{fig:widths}. 

More formally, we introduce:

\vspace{-1em}
\begin{equation}
\omega_{pqhwk}(\theta) = \exp(-\frac{(h-\alpha_{pk}(\theta))^2+(w-\beta_{qk}(\theta))^2}{2\sigma^2}) \\
\label{eq:cond_diff}
\end{equation} 
\begin{equation*}
\text{where } \begin{cases}
    \alpha_{pk}(\theta) = p\cdot str_h - (k-pad)\cdot \sin \theta \\
\beta_{qk}(\theta) = q\cdot str_w + (k-pad)\cdot \cos \theta \\ \end{cases}
\end{equation*} 
\vspace{-1em}

Modulo discretization, this formulation generalizes \Cref{eq:sum_noomega}. Indeed, it can be seen as \Cref{eq:sum_omega} for the special case $\sigma = 0$ or in other words
$$\omega_{pqhwk}(\theta)=\mathds{1}(f_{p,k}(\theta)=h) \mathds{1}(g_{q,k}(\theta)=w)$$
\vspace{-0.5em}

By introducing \Cref{eq:sum_omega} and our parameterization (\ref{eq:cond_diff}) of $\omega$, we have described a differentiable formulation of \Cref{eq:sum_noomega} which is differentiable w.r.t $\theta$. 

Intuitively, $\sigma$ models the error in constraint \Cref{eqn:coordinate1d} and consequently how much we want neighboring pixels to influence the output of the convolution. By varying $\sigma$ we control the width of this band as well as the pixels that have an influence over the optimization of $\theta$. 

However, by increasing $\sigma$, we sacrifice speed: the computational cost rises from $O(K)$ to $O(HWK)$. This can be offset to $O(r^2K)$ by leveraging the exponential decay of $\omega$ to cut off contributions below a threshold $\exp(-\frac{r^2}{2\sigma^2})$. $r$ can be seen as the radius by which we expand the kernel.

Optimizing $\theta$ is beyond the scope of the paper and this section is provided only for reference. It can be extended to support per-channel angles $\boldsymbol \theta$ in a similar way. We leave exploration of this idea as future work.

\section{Connection with 2D Gaussian anisotropic filters}
\label{anisotropic_filters}

\vspace{-0.25em} 

In this section, we make a parallel between oriented 1D kernels and gaussian anisotropic filters.
Before going further, let's first introduce 2D gaussian filters, which are commonly  used in signal processing \cite{canny_edge_detector} as:

\vspace{-0.25em}

\begin{definition}
A 2D gaussian filter $k(\mathbf{x},\mathbf{y})$ with positive semi-definite covariance matrix $\boldsymbol \Sigma \! \in \! \mathbb{R}^{2\times 2}$ is defined as:
\vspace{-0.5em}
\begin{align}
    k(\mathbf{x}, \mathbf{y}) = \frac{1}{2\pi |\boldsymbol\Sigma|^\frac{1}{2}}\exp (-\frac{1}{2}(\mathbf{x}-\mathbf{y})^T \boldsymbol\Sigma^{-1} (\mathbf{x}-\mathbf{y}))
\end{align}

\vspace{-1em}
for given $\mathbf{x}, \mathbf{y} \in \mathbb{R}^2$.

\begin{enumerate}
    \item \vspace{-0.25em} In the case where $\boldsymbol\Sigma = \sigma \mathbf{I}_2$ for a given $\sigma \geq 0$, we say that the filter is \emph{isotropic}.
    \item \vspace{-0.25em} In the case where $\boldsymbol\Sigma = \text{Diag}(\sigma_1, \sigma_2)$ we say that the filter is \emph{orthogonal}.
    \item \vspace{-0.25em} Otherwise, the filter is said to be \emph{anisotropic}.
\end{enumerate}
\vspace{-0.25em} 

\end{definition}

\begin{figure}[h]
    \centering
    \includegraphics[width=\linewidth]{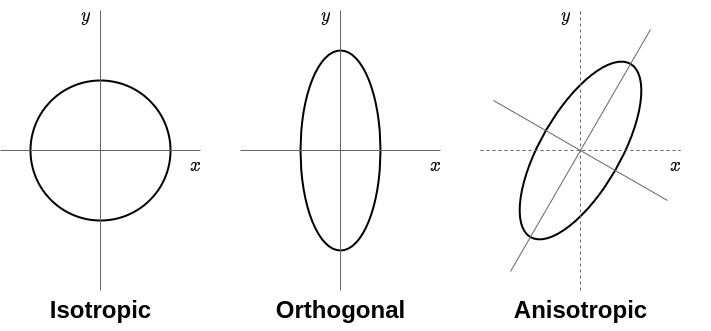}
    \vspace{-2em}
    \caption{Examples of 2D gaussian filters.}
    \vspace{-0.5em} 
    \label{fig:convnext}
\end{figure}

\subsection{Parallel with gaussian anisotropic filters}

\vspace{-0.25em} 

In gaussian filtering literature \cite{geusebroek_gaussian_2003}, it is commonly stated that 2D isotropic and orthogonal gaussian filters are separable and can be decomposed as the recursion of a horizontal 1D gaussian filter and of a vertical 1D gaussian filter. In practice, gaussian filters are used in the form of finite-sized convolutions. This means that 2D isotropic/orthogonal kernels can be decomposed as the recursion of a horizontal and vertical convolution.

According to \cite{geusebroek_gaussian_2003, lampert_gaussian_2006}, this fact is not limited to ``axis-aligned" gaussian filters: any 2D gaussian filter can be expressed as the recursion of two \textit{oriented 1D} gaussian filters, as dictated by the eigenvectors of its covariance matrix. Consequently, the combination of two oriented 1D convolutions can represent \textit{any} 2D gaussian filter including anisotropic ones, thereby suggesting that oriented 1D kernels are more expressive than non-oriented kernels. The situation is summarized in \Cref{tab:oriented_anistropic}.

\begin{table}[h]
    \centering
    \begin{tabular}{l|c|c|c}
         Convolution kernel  & Isotropic & Orthogonal & Anisotropic \\
         \hline 
         Non-oriented & \textbf{Yes} & \textbf{Yes} & \textit{No} \\
         Oriented & \textbf{Yes} & \textbf{Yes} & \textbf{Yes}
    \end{tabular}
    \caption{Summary of the expressiveness of oriented and non-oriented convolution kernels}
    \vspace{-1em} 
    \label{tab:oriented_anistropic}
\end{table}

\vspace{1em}

\newpage

\section{Implementation Notes}
\label{implementation_notes}

\vspace{-0.25em} 

In this section, we look at the challenges involved in implementing fast depthwise convolutions for oriented 1D kernels. We present here two main approaches to implement oriented kernels: the \textit{filter rotation} approach, which rotates the kernel, and the \textit{input rotation} approach, which keeps the kernel fixed and instead rotates the input in the opposite direction.  Our best algorithm runs up to $50\%$ faster than PyTorch on 1 NVIDIA RTX 3090, as presented in the paper.

\vspace{-0.25em}

\subsection{CUTLASS GEMM with Rotated Filter}

\vspace{-0.25em}

Our goal is to design an implementation that works for large kernel sizes $K \geq 7$. To that effect, our first proposed implementation extends an open-source library that achieves state-of-the-art performance on 2D depthwise convolution, namely \textit{MegEngine Cutlass}\cite{ding_scaling_2022} \footnote{\url{https://github.com/MegEngine/cutlass}}. It leverages specialized GEneral Matrix Multiplication (GEMM) primitives that are known to be very efficient \cite{markidis_nvidia_2018}. The codebase was initially introduced in \cite{ding_scaling_2022} and forked from \textit{NVIDIA Cutlass}\cite{cutlass} \footnote{\url{https://github.com/NVIDIA/cutlass}}. 

The original non-oriented implementation computes the offsets $(h,w)$ by linearly increasing $k$, thereby achieving contiguous memory reads. In this proposed implementation of oriented kernels, we instead compute $(h,w)$ using \Cref{eqn:coordinate1d} which unfortunately introduces non-contiguous memory read issues.

Using this proposed implementation results in uneven speeds with respect to a given angle $\theta$: 45$^\circ$ angles are more than 2$\times$ as slow compared to horizontal kernels (see \Cref{table:speed}).

\vspace{-0.25em}

\subsection{CUTLASS GEMM with Rotated Input}

\vspace{-0.25em}

As an alternative to rotating the filter, we can instead keep the filter fixed, and rotate the input in the opposite direction. The selling point of this approach is that we can use existing optimized depthwise convolution implementations without modification. The downside is that we introduce extra overhead by adding input and output image rotation steps. Furthermore, to preserve the information content, the rotations grow the image size $HW$ up to 2$\times$ its original value, which leads to slower convolutions. We consider additional optimization tricks for this approach such as image compression. 

\begin{figure}[h]
     \centering
     \raisebox{1em}{\begin{subfigure}[b]{0.3\linewidth}
         \centering
         \includegraphics[scale=0.5]{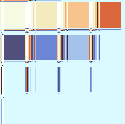}
         \caption{Rotated filter convolution}
         \label{fig:rotated_input_impl}
     \end{subfigure}}
     \hfill
     \begin{subfigure}[b]{0.3\linewidth}
         \centering
         \includegraphics[scale=0.5]{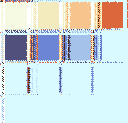}
         \caption{Rotated input convolution: Nearest neighbor}
         \label{fig:rotated_input_compressed}
     \end{subfigure}
     \hfill
     \begin{subfigure}[b]{0.3\linewidth}
         \centering
         \includegraphics[scale=0.5]{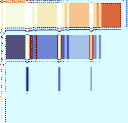}
         \caption{Rotated input convolution: Bilinear interpolation}
         \label{fig:rotated_input_compressed}
     \end{subfigure}
     \vspace{-1em}
     \caption{Qualitative analysis of aliasing in the rotated input approach. Input is a tiled square with tiles of varying color. Even with bilinear interpolation, border error is significant, which becomes a problem for real-world inputs. }
     \vspace{-1em}
    \label{fig:aliasing}
\end{figure}

One of the drawbacks of the rotated input approach is that it introduces \textit{aliasing} due to image rotation as shown in \Cref{fig:aliasing}. For small image sizes, this greatly perturbs the output and leads to significant drops in accuracy. We considered several interpolation schemes (bilinear, lanczos \cite{turkowski_filters_1990}, ...) to account for aliasing but they do not help with accuracy. 

\subsection{CUDA from scratch}

\vspace{-0.25em}

Our best performing algorithm is a custom CUDA kernel implemented from scratch. The idea of this algorithm stems from the observation that 1D depthwise convolutions incur a low computation-to-bandwidth ratio: in other words, optimizing data access pattern is critical in achieving good performances. 
Optimizing these data patterns, we obtain an implementation that is even faster compared to non-oriented horizontal PyTorch\cite{pytorch}/CuDNN\cite{cudnn} convolutions.

We accomplish this by loading the whole input in shared GPU memory before starting our computations. This allows us to avoid the costly non-coalesced global memory accesses caused by the oriented nature of our computations.
However, for larger image sizes, the input does not fit into shared GPU memory. As a result, we choose to cut the image into vertical bands and compute oriented 1D convolutions on each band. We also attempt to maximize data sharing by splitting the data loads between GPU threads so that every pixel of the input is read only once (if we ignore  bordering bands). We further optimize data access by adopting vectorized loads, which greatly increase data throughput. Finally, we find the set of parameters that maximize performance on a given hardware, and achieve better performance compared to PyTorch, even though we support arbitrary oriented 1D kernels. Note that this implementation requires a specific CUDA kernel for every input size.

\vspace{-0.25em}

\subsection{Rotated Input Compression}
\vspace{-0.25em} 

\begin{figure}
     \centering
     \begin{subfigure}[b]{0.4\linewidth}
         \centering
         \fbox{\includegraphics[scale=0.2]{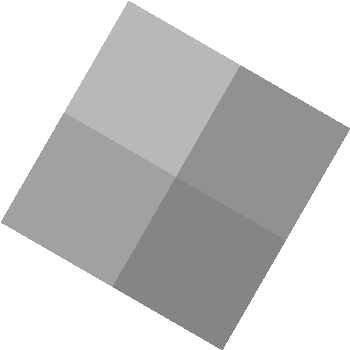}}
         \caption{Rotated input}
         \label{fig:rotated_input_impl}
     \end{subfigure}
     \hfill
     \raisebox{1.3em}{\begin{subfigure}[b]{0.4\linewidth}
         \centering
         \fbox{\includegraphics[scale=0.2]{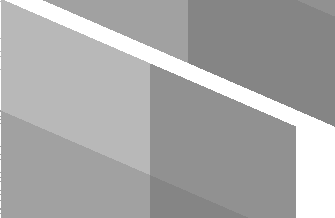}}
         \caption{Compressed input}
         \label{fig:rotated_input_compressed}
     \end{subfigure}}
     \vspace{-0.25em} 
     \caption{Visualization of a tiled square rotated by our rotated input kernel. After rotation, the input is fed to a standard non-oriented horizontal 1D kernel and the output is rotated back to obtain the final result. On the right, we show how compression intuitively splits the input before passing it to the horizontal kernel - the speedup comes from reducing the input size and avoiding unnecessary computations. }
     \vspace{-0.5em} 
    \label{fig:rotated_input}
\end{figure}

From a performance perspective, the \textit{input rotation} approach suffers from two inefficiencies: 1) it introduces additional pre- and post-convolution steps, which can increase runtime by up to 40\% compared to a non-oriented horizontal convolution (see \Cref{table:speed}), and 2) the image rotation can grow the intermediate image size to 2$\times$ the original size, as shown in \Cref{fig:rotated_input_impl}. This means that the convolution can be 2$\times$ as costly. 

To mitigate issue 2), we can compress the rotated image in a way that does not affect the convolution operation as shown in \Cref{fig:rotated_input_compressed}. Intuitively, we compress together the tips of the square in order to reduce $H$, and we align the sides of the square with the left border to reduce $W$.
In theory, we can expect to reduce the image size to its original value, because area is preserved through rotation.  We add a padding between the two blocks to preserve the result of the convolution. This increases the compressed size $H'W'$ by $KW'$.

\begin{table*}[t]
\scriptsize
\centering
\begin{tabular}{ r|cccccccc|cccccccc } 
 \hline
 $K$  & \multicolumn{8}{c|}{7} & \multicolumn{8}{c}{31} \\
 \hline
 Angle & 0 & 22.5 & 45 & 67.5 & 90 & 112.5 & 135 & 157.5 & 0 & 22.5 & 45 & 67.5 & 90 & 112.5 & 135 & 157.5 \\
 \hline
 \rowcolor{gray!10} & \multicolumn{16}{c}{$H=W=14$} \\
 \hline
 PyTorch/CuDNN Horiz. only & \multicolumn{8}{c|}{0.4} & \multicolumn{8}{c}{1.1}\\
 CUTLASS Horiz. only \cite{ding_scaling_2022} & \multicolumn{8}{c|}{0.6} & \multicolumn{8}{c}{0.6} \\
 CUTLASS Rotated Filter (Ours) & 0.7 & 0.7 & 0.8 & 0.8 & 0.8 & 0.6 & 0.6 & 0.7 & 0.7 & 0.8 & 0.8 & 0.8 & 0.8 & 0.7 & 0.6 & 0.8\\
 CUTLASS Rotated Input (Ours) & 0.9 & 1.5 & 1.5 & 1.5 & 1.4 & 1.5 & 1.0 & 1.4 & 0.9 & 1.6 & 1.6 & 1.6 & 1.4 & 1.6 & 1.0 & 1.4\\
 CUDA from scratch (Ours) & 0.3 & 0.3 & 0.3 & 0.3 & 0.3 & 0.3 & 0.3 & 0.3 & 0.8 & 0.8 & 0.8 & 0.7 & 0.7 & 0.8 & 0.8 & 0.8\\
 \hline
 \rowcolor{gray!10} & \multicolumn{16}{c}{$H=W=28$} \\
 \hline
 PyTorch/CuDNN Horiz. only & \multicolumn{8}{c|}{1.4} & \multicolumn{8}{c}{3.8}\\
 CUTLASS Horiz. only \cite{ding_scaling_2022} & \multicolumn{8}{c|}{2.3} & \multicolumn{8}{c}{2.5} \\
 CUTLASS Rotated Filter (Ours) & 3.1 & 5.3 & 6.2 & 6.0 & 6.2 & 5.1 & 2.5 & 5.1 & 3.1 & 6.8 & 7.8 & 7.6 & 7.8 & 6.6 & 2.7 & 6.4\\
 CUTLASS Rotated Input (Ours) & 3.3 & 5.7 & 5.7 & 5.9 & 5.6 & 5.7 & 4.0 & 5.2 & 3.5 & 6.1 & 6.0 & 6.2 & 5.9 & 6.0 & 4.2 & 5.5\\
 CUDA from scratch (Ours) & 0.9 & 0.9 & 0.9 & 0.9 & 1.0 & 0.9 & 0.9 & 0.9 & 2.6 & 2.6 & 2.6 & 2.6 & 2.5 & 2.6 & 2.6 & 2.6\\
 \hline
 \rowcolor{gray!10} & \multicolumn{16}{c}{$H=W=56$} \\
 \hline
 PyTorch/CuDNN Horiz. only & \multicolumn{8}{c|}{5.5} & \multicolumn{8}{c}{15.0}\\
 CUTLASS Horiz. only \cite{ding_scaling_2022} & \multicolumn{8}{c|}{9.4} & \multicolumn{8}{c}{10.2} \\
 CUTLASS Rotated Filter (Ours) & 13.4 & 49.6 & 61.7 & 60.4 & 61.8 & 52.9 & 14.8 & 46.1 & 13.6 & 66.1 & 85.3 & 81.5 & 86.0 & 70.3 & 17.7 & 60.3\\
 CUTLASS Rotated Input (Ours) & 13.3 & 22.4 & 22.7 & 23.4 & 21.6 & 22.6 & 15.6 & 20.7 & 14.1 & 24.1 & 24.4 & 24.7 & 23.1 & 24.3 & 16.6 & 22.2\\
 CUDA from scratch (Ours) & 3.2 & 3.3 & 3.2 & 3.2 & 3.2 & 3.2 & 3.2 & 3.2 & 9.8 & 9.8 & 9.8 & 9.9 & 9.9 & 9.9 & 10.0 & 10.0\\
 \hline
\end{tabular}
\caption{Full runtime comparison of the different oriented 1D convolution implementations on an NVIDIA RTX 3090 for $N=64, C=512$, FP32. The mean is taken over 100 runs, preceded by 10 dry runs. We benchmark against the very competitive CuDNN/PyTorch and CUTLASS implementations. Our \textit{CUDA from scratch} outperforms consistently all other implementations regardless of angle with intelligent data access patterns, but starts to fall off when computation becomes the bottleneck. Our CUTLASS based methods either rotate the filter, which leads to non-contiguous/slow memory accesses; or either rotate the input, which enable the use of the efficient horizontal CUTLASS kernel but introduce aliasing and grow the image size $HW$ up to 2$\times$. }
\vspace{-2em}
\label{table:speed}
\end{table*}

In practice, this compression improves overall speed. However, it requires careful implementation as it can slow down the rotation steps quite significantly. 

\vspace{-0.25em}

\subsection{Benchmarks}

\vspace{-0.25em}

\Cref{table:speed} compares training speeds for our 1D oriented kernel implementations. \Cref{tab:speed2} presents a more compact summary, and shows training speed as well as inference speed.  Our \textit{CUDA from scratch} beats consistently all other implementations regardless of angle with intelligent data access patterns, but starts to fall off when computation becomes the bottleneck. Our CUTLASS based methods either rotate the filter, which leads to non-contiguous/slow memory accesses; or either rotate the input, which enable the use of the efficient horizontal CUTLASS kernel but introduce aliasing and grow the image size $HW$ up to 2$\times$.

We complement these kernel-level benchmarks with network-level benchmarks in \Cref{tab:network_speed}. It compares training and inference throughputs measured in img/s of ConvNeXt-based models versus RepLKNet and SLaK, on an input of size $224^2$. 

In this paper, we focused on Depthwise Separable Convolutions (DSCs) as most modern ConvNets combine depthwise and pointwise together. Note that this setting is more challenging as improving the performance of depthwise convolutions does not necessarily lead to better DSC performance. 
If we remove this assumption, our oriented 1D depthwise convolutions are clearly faster than 2D depthwise convolutions, in theory and practice. We show this for large kernels in \Cref{table:computational_analysis} of our paper, and add benchmarks in \Cref{table:table_2D_1D}. 
\vspace{-0.5em}

\begin{table}[h]
\small
    \centering
    \begin{tabular}{l|cccc}
         Implementation & $K$ & Angle & Inference & Training \\ 
         & & & Time & Time \\
         \hline 
         PyTorch Horiz. & 31 & 0$^\circ$ & 5.2$\pm$0.1ms & 14.9$\pm$0.3ms \\
         CUTLASS Horiz. & 31 & 0$^\circ$ & 3.4$\pm$0.1ms & 10.1$\pm$0.1ms \\
         \hline 
         CUDA From Scratch & 31 & 0$^\circ$ & 4.2$\pm$0.1ms & 9.9$\pm$0.1ms\\
         CUDA From Scratch & 31 & 45$^\circ$ & 4.2$\pm$0.1ms & 9.9$\pm$0.1ms\\
         Rotated Filter & 31 & 0$^\circ$ & 5.4$\pm$0.1ms & 13.6$\pm$0.1ms\\
         Rotated Filter & 31 & 45$^\circ$ & 30.7$\pm$0.2ms & 85.3$\pm$0.3ms\\
         Rotated Input & 31 & 0$^\circ$ & 5.3$\pm$0.1ms & 14.1$\pm$0.1ms\\
         Rotated Input & 31 & 45$^\circ$ & 9.4$\pm$0.1ms & 24.6$\pm$0.1ms\\
         Rotated Compressed & 31 & 0$^\circ$ & 5.3$\pm$0.1ms & 13.9$\pm$0.1ms \\
         Rotated Compressed & 31 & 45$^\circ$ & 4.8$\pm$0.1ms & 13.0$\pm$0.1ms
    \end{tabular}
    \vspace{-0.25em}
    \caption{\footnotesize Comparison of our different implementations, for $N=64, C=512, H=W=56$ on an NVIDIA RTX 3090, PyTorch 1.11, CUDA 11.3, CuDNN 8.2, FP32. We see that Rotated Compressed improves performance for non-zero angles. 
    \textit{Inference Time} measures only forward pass, \textit{Training Time} includes backpropagation. }
    \vspace{-0.5em}
    \label{tab:speed2}
\end{table}

\begin{table}[h]
    \vspace{-0.5em}
\footnotesize
    \centering
    \begin{tabular}{l|cc}
         Model & Inference & Training \\
         & Throughput & Throughput \\
         \hline 
         ConvNeXt-B & 460 & 140\\
         ConvNeXt-1D-B & 450 & 140\\
         RepLKNet-B & 340 & 90 \\
         ConvNeXt-1D++-B & 290 & 105\\
         ConvNeXt-2D++-B & 290 & 100\\
         SLaK-B & 210 & 60\\
         
    \end{tabular}
    \vspace{-0.25em}
    \caption{\small Benchmark of ConvNeXt, RepLKNet and SLaK on a $32\times 224^2$ FP32 batch. 
    \textit{Inference Throughput} measures the number of images per second, forward pass only, \textit{Training Throughput} includes backpropagation. }
    \vspace{-0.5em}
    \label{tab:network_speed}
\end{table}

\begin{table}
\scriptsize
\hspace{+1em}
\begin{tabular}{l|ccc}
 \hline
 Kernel & $K$ & Inference time & Train time \\
 \hline
 PyTorch 2D & $31 \times 31$ & 110ms & 410ms \\
 PyTorch 1D & $1 \times 31$ & 5.2ms & 14.9ms \\ 
 Ours 1D & $1 \times 31$ & 4.2ms & 14.7ms \\ 
 \hline
 PyTorch 2D & $15 \times 15$ & 29ms & 95ms\\
 PyTorch 1D & $1 \times 15$ & 3.4ms & 8.7ms\\ 
 Ours 1D & $1 \times 15$ & 2.0ms & 7.5ms\\ 
 \hline
 PyTorch 2D & $7 \times 7$ & 8.2ms & 22ms\\
 PyTorch 1D & $1 \times 7$ & 2.4ms & 5.4ms\\ 
 Ours 1D & $1 \times 7$ & 1.0ms & 4.2ms\\ 
 \hline
 PyTorch 2D & $3 \times 3$ & 2.2ms & 5.8ms\\ 
 PyTorch 1D & $1 \times 3$ & 2.0ms & 4.0ms\\ 
 Ours 1D & $1 \times 3$ & 1.0ms & 3.9ms\\
 \hline
\end{tabular}
\vspace{-0.25em}
\caption{\small Comparison between PyTorch 1D, 2D and our oriented kernels on an input of batch size 64, $C\! =\! 512$ and $H\! =\! W\! =\! 56$, FP32, measured on 1 NVIDIA RTX3090. We see that 1D kernels are clearly faster than 2D kernels of same kernel size and that our implementation is able to match PyTorch efficiency. \textit{Inference} measures forward pass, \textit{Training} includes backpropagation.}
\vspace{-0.5em}
\label{table:table_2D_1D}
\end{table}

\newpage

\section{Training Settings}
\label{training_settings}
\vspace{-0.25em} 

\subsection{Pre-training}

\vspace{-0.25em} 

In this section, we provide the full training settings used in our experiments. We adapt them directly from ConvNeXt \cite{liu_convnet_2022}. All settings apart from stochastic depth rate are the same for all model variants, as described in \cite{liu_convnet_2022} and are listed in \Cref{tab:full_training_settings}.

\begin{table}[h]
    \centering
    \begin{tabular}{l|c}
         Setting & ConvNeXt-1D/1D++/2D++ \\
         & T/S/B \\
         \hline
         weight init & trunc. normal (0.2)\\
optimizer & AdamW \\
base learning rate & 4e-3 \\
weight decay & 0.05 \\
optimizer momentum & $\beta_1, \beta_2=0.9, 0.999$ \\
batch size & 4096 \\
training epochs & 300 \\
learning rate schedule & cosine decay \\
warmup epochs & 20 \\
warmup schedule & linear \\
layer-wise lr decay \cite{bao2022beit, clark2020electra} & None \\
randaugment \cite{cubuk_randaugment_2019} & (9, 0.5) \\
mixup \cite{zhang_mixup_2018} & 0.8 \\
cutmix \cite{yun_cutmix_2019} & 1.0 \\
random erasing \cite{zhong_random_2020} & 0.25 \\
label smoothing \cite{szegedy_rethinking_2016} & 0.1 \\
stochastic depth \cite{huang_deep_2016} & 0.1/0.4/0.5 \\
layer scale \cite{touvron_going_2021} & 1e-6 \\
head init scale \cite{touvron_going_2021} & None \\
gradient clip & None \\
exp. mov. avg. (EMA) \cite{polyak_ema_1992} & 0.9999 \\
    \end{tabular}
    \caption{\textbf{ImageNet training settings.} Input is of size 224$^2$. Settings taken from ConvNeXt \cite{liu_convnet_2022}. Stochastic depth rates 0.1/0.4/0.5 are dependent on model size T/S/B.}
    \vspace{-1em} 
    \label{tab:full_training_settings}
\end{table}

\subsection{Downstream tasks}

For ADE20K and COCO experiments, we follow the same settings as ConvNeXt, and use the same toolboxes MMDetection \cite{mmdetection} and MMSegmentation \cite{mmsegmentation}. We also use non-EMA weights. For COCO experiments, we train a Cascade Mask-RCNN \cite{cascade_maskrcnn} network for 36 epochs, with a 3$\times$ schedule, a learning rate of 2e-4, a layer-wise l.r. decay of 0.7/0.8 and a stochastic depth rate of 0.4/0.7 for \textit{Tiny}/\textit{Base} respectively. For ADE20K, we train a Upernet \cite{upernet} network, for 160k iterations, with a learning rate of 1e-4, layer-wise l.r. decay of 0.9,  stochastic depth rate of 0.4 and report validation mIoU results using multi-scale testing.

\section{Limitations and Future Work}
\label{limitations}
\vspace{-0.5em}

As mentioned in previous sections, some aspects of oriented 1D kernels deserve more exploration. This includes looking at oriented non-depthwise convolutions, and doing angle backpropagation. We have mae a lot of design choices which constrain 1D kernels and their expressiveness. As future work, we would like to relax these hypotheses and come up with efficient implementations for more general cases. 
In this paper, we have demonstrated that we can expect measurable benefits by using oriented 1D kernels, which shows how promising such an approach can be if explored further.

\newpage

{\small
\bibliographystyle{ieee_fullname}
\bibliography{references}
}

\end{document}